\documentclass[sigconf,10pt]{acmart}

\makeatletter
\def\@ACM@checkaffil{%
    \if@ACM@instpresent\else
    \ClassWarningNoLine{\@classname}{No institution present for an affiliation}%
    \fi
    \if@ACM@citypresent\else
    \ClassWarningNoLine{\@classname}{No city present for an affiliation}%
    \fi
    \if@ACM@countrypresent\else
        \ClassWarningNoLine{\@classname}{No country present for an affiliation}%
    \fi
}
\makeatother

\acmYear{2023}\copyrightyear{2023}
\setcopyright{acmlicensed}
\acmConference[ACM MobiCom '23]{The 27th Annual International Conference on Mobile Computing and Networking}{October 2--6, 2023}{Madrid, Spain}
\acmBooktitle{The 27th Annual International Conference on Mobile Computing and Networking (ACM MobiCom '23), October 2--6, 2023, Madrid, Spain}
\acmPrice{15.00}
\acmDOI{10.1145/3570361.3592532}
\acmISBN{978-1-4503-9990-6/23/10}

\usepackage{epsfig}
\usepackage{subcaption}
\usepackage{graphicx}
\usepackage{amsmath}
\usepackage{float}
\usepackage[font=small,labelfont=bf]{caption}
\usepackage{wrapfig}
\usepackage{url}
\usepackage{hyperref}

\usepackage{breakurl}
\usepackage{tabularx, multirow, multicol, booktabs}
\usepackage{caption}
\usepackage{graphbox}
\usepackage{lipsum}
\usepackage[linesnumbered,algoruled,boxed,lined]{algorithm2e}
\usepackage{xcolor}
\usepackage{ulem}
\usepackage{titlesec}

\titlespacing\subsection{0pt}{6pt plus 2pt minus 2pt}{6pt plus 0pt minus 4pt}
\titlespacing\subsubsection{0pt}{6pt plus 2pt minus 2pt}{6pt plus 0pt minus 4pt}

\let\svthefootnote\thefootnote
\newcommand\freefootnote[1]{%
  \let\thefootnote\relax%
  \footnotetext{#1}%
  \let\thefootnote\svthefootnote%
}

\makeatletter 
\g@addto@macro{\@algocf@init}{\SetKwInOut{Parameter}{Parameters}} 
\makeatother
\SetKwInput{KwInput}{Input}                %
\SetKwInput{KwOutput}{Output}              %

\newcommand{\name}{BatMobility}
\newcommand{\para}[1]{\vspace{4pt}\noindent\textbf{#1}}
\newcommand{\squishlist}
{
    \begin{list}{$\bullet$}
    {
        \setlength{\itemsep}{0pt}      \setlength{\parsep}{3pt}
        \setlength{\topsep}{3pt}       \setlength{\partopsep}{0pt}
        \setlength{\leftmargin}{1.5em} \setlength{\labelwidth}{1em}
        \setlength{\labelsep}{0.5em}
    }
}
\newcommand{\squishend}
{
    \end{list}
}

\newcommand{\enumend}
{
    \end{enumerate}
}
\newcommand{\squishenum}
{
    \begin{enumerate}
    {
        \setlength{\itemsep}{0pt}      \setlength{\parsep}{3pt}
        \setlength{\topsep}{3pt}       \setlength{\partopsep}{0pt}
        \setlength{\leftmargin}{1.5em} \setlength{\labelwidth}{1em}
        \setlength{\labelsep}{0.5em}
    }
}

\newcommand{\red}[1]{\textcolor{black}{#1}}
\newcommand{\textred}[1]{\textcolor{black}{#1}}

\AtBeginDocument{%
  \providecommand\BibTeX{{%
    \normalfont B\kern-0.5em{\scshape i\kern-0.25em b}\kern-0.8em\TeX}}}

\begin{document}

\title{BatMobility: Towards Flying Without Seeing for Autonomous Drones}

\author{Emerson Sie, Zikun Liu, and Deepak Vasisht} 
\affiliation{ 
    \institution{University of Illinois Urbana-Champaign}
    \vspace{-0.1in}
}

\begin{abstract}
Unmanned aerial vehicles (UAVs) rely on optical sensors such as cameras and lidar for autonomous operation. However, such optical sensors are error-prone in bad lighting, inclement weather conditions including fog and smoke, and around textureless or transparent surfaces. In this paper, we ask: is it possible to fly UAVs without relying on optical sensors, i.e., can UAVs fly without seeing? We present \name, a lightweight mmWave radar-only perception system for UAVs that eliminates the need for optical sensors. \name\ enables two core functionalities for UAVs -- radio flow estimation (a novel FMCW radar-based alternative for optical flow based on surface-parallel doppler shift) and radar-based collision avoidance. We build \name\ using commodity sensors and deploy it as a real-time system on a small off-the-shelf quadcopter running an unmodified flight controller. Our evaluation\footnotemark[1] shows that \name\ achieves comparable or better performance than commercial-grade optical sensors across a wide range of scenarios.
\color{black}
\end{abstract}

\ccsdesc[500]{Computer systems organization~Robotics}
\ccsdesc[500]{Computing methodologies~Machine learning}
\ccsdesc[500]{Computer systems organization~Sensors and actuators}

\keywords{Machine Learning, Quadrotor, Radar, Egomotion, RF Sensing}

\maketitle

\section{Introduction}\label{sec:intro}
\footnotetext[1]{\label{weblink}Videos are available at the project website: \href{https://batmobility.github.io}{\color{blue}{https://batmobility.github.io}}}

\begin{figure}
    \includegraphics[width=\linewidth]{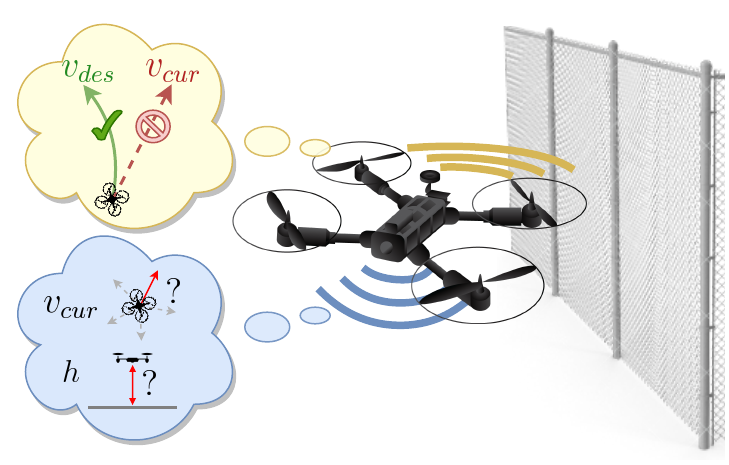}
    \caption{A downward facing radar performs altimetry and ego-velocity estimation (bottom) and a forward facing radar performs collision avoidance (top), removing the need for other exteroceptive sensors such as cameras or lidar.}\vspace{-0.1in}
    \label{fig:intro}
\end{figure}

\begin{figure*}
    \centering
    \begin{subfigure}{\textwidth}
    \centering
    \includegraphics[width=0.28\textwidth]{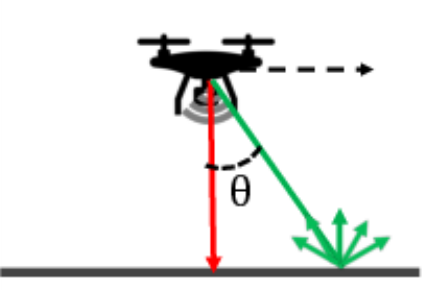}
    \includegraphics[width=0.32\textwidth]{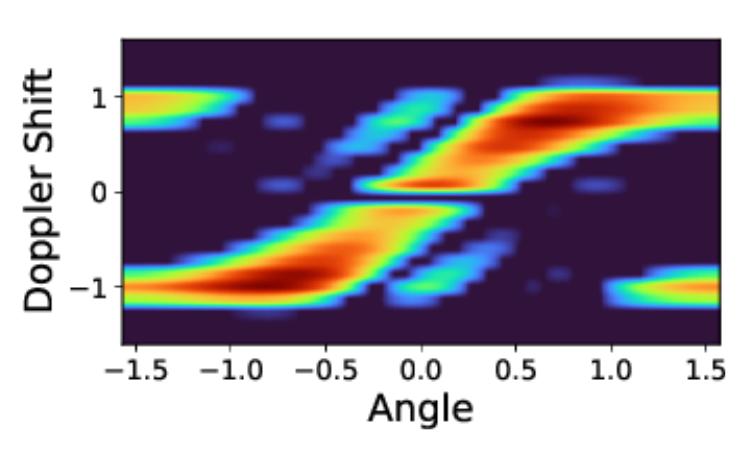}
    \includegraphics[width=0.32\textwidth]{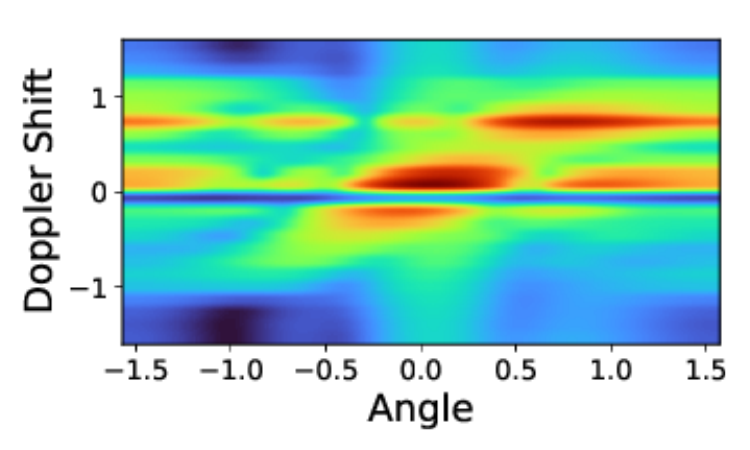}\vspace{-0.1in}
    \caption{(left) Motion of the UAV, (center) Simulated doppler-angle plot, (right) Observed doppler-angle plot.}
    \end{subfigure}
    \begin{subfigure}{\textwidth}
    \centering
    \includegraphics[width=0.28\textwidth]{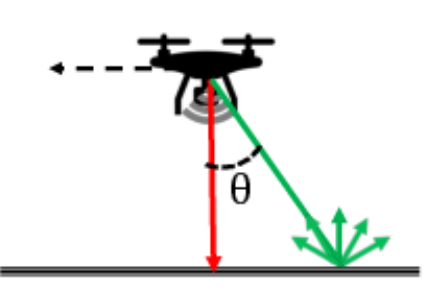}
    \includegraphics[width=0.32\textwidth]{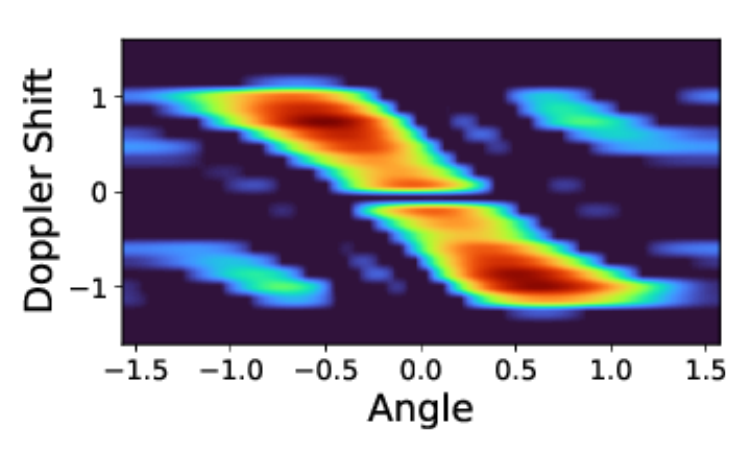}
    \includegraphics[width=0.32\textwidth]{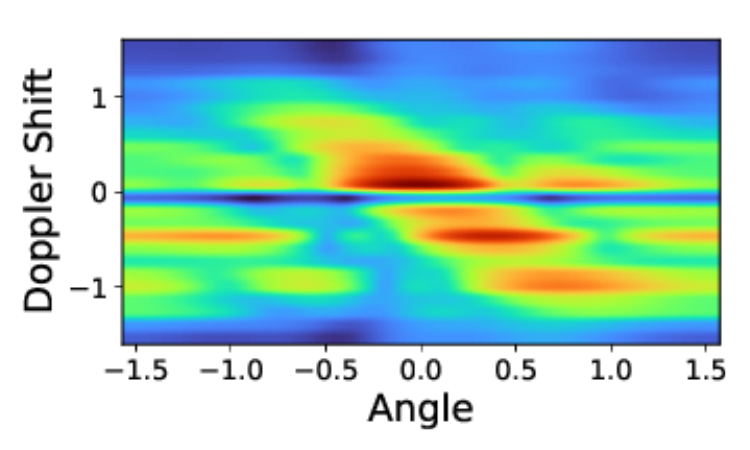}\vspace{-0.1in}
    \caption{(left) Motion of the UAV, (center) Simulated doppler-angle plot, (right) Observed doppler-angle plot.}\vspace{-0.1in}
    \end{subfigure}
    \caption{Estimating Surface-Parallel Doppler Shifts: At mmWave frequencies, even seemingly flat surfaces scatter the signal (green arrows) leading to continuous signatures in our doppler-angle plots. Observed signals are impacted by additional factors: clutter, multipath, and noise. }%
    \label{fig:da_sim_vs_real}
\end{figure*}

In recent years, we have experienced a proliferation of small UAVs meant for many diverse applications like inventory management in warehouses \cite{ma_drone_2017}, cargo delivery \cite{rwanda, ukraine}, building inspections \cite{inspection, uav_sensing3, uav_auto3}, mapping \& surveying \cite{uav_sensing1, uav_sensing2}, home surveillance \cite{ring}, and search \& rescue operations \cite{rescue}. \red{Due to their wide ranging applications,} the market size for small UAVs is expected to increase to nearly 50 billion US dollars by 2028. \red{However, most UAVs still rely on manual teleoperation by humans.} This has led to much research in sensing~\cite{uav_sensing1, uav_sensing2, uav_sensing3}, automation~\cite{uav_auto1, uav_auto2, uav_auto3}, and control~\cite{uav_con1, uav_con2, uav_con3} for UAVs.

\textred{Current approaches for autonomous UAVs mainly rely on optical sensors such as cameras or lidars to perceive the surrounding environment. Systems based on optical sensing readily benefit from decades of computer vision research, but suffer from multiple shortcomings.} First, optical sensors are prone to failures in visually degraded conditions due to, e.g., bad lighting conditions, inclement weather conditions like fog, and \textred{occlusions like dirt}. Second, optical sensors fail to identify relative motion in untextured or featureless environments, e.g., large warehouse floors due to the lack of visual \red{or geometric} features. Similarly, \red{optical sensors fail to register} transparent objects like glass windows and panels. Finally, and perhaps less obviously, optical sensors raise privacy concerns, such as in the case of outdoor delivery UAVs flying over residential areas \cite{drone_survey}.

\textred{Radio-frequency (RF) sensing has emerged as an alternative paradigm to optical sensing in recent years \cite{adib_see_2013, adib_3d_2014, zhao_rf-based_2018, guan_through_2020, bansal_pointillism_2020}. Given the progress in RF sensing,} we ask a simple yet seemingly radical question: \textit{is it possible for UAVs to operate autonomously without using optical sensors at all?} In other words, \textit{can UAVs autonomously fly without seeing?}

\textred{To answer this question, we explore the use of commodity single-chip mmWave radars as the sole perception tool on autonomous UAVs.} Such radars offer several advantages not only over optical sensors, but over all other types of sensors for UAVs (i.e., ultrasound) as well. First, they are invariant to light levels \textred{allowing for 24/7 operation}, are robust to weather-related precipitates like fog, and can operate over long range. Second, they only collect coarse-grained information about the surrounding environment and alleviate privacy concerns. Third, they are compact in terms of size and power consumption, \textred{contain no moving parts}, and can furthermore be hidden behind a facade to prevent the need for frequent cleaning due to dirt and grime. In spite of these advantages, radar is used for limited operations on UAVs (e.g., estimating height from ground) and not as a first-class perception tool.

\textred{We introduce \name, a new perception system that relies purely on inexpensive radar, does not have any optical sensors, and can run on small and computationally constrained UAVs}.  Our contribution in this paper is two-fold. First, we use radar to enable two core navigation primitives that have traditionally been implemented using optical sensors: \textred{(a) estimating an UAV's velocity and altitude with respect to the ground to stabilize a UAV and to enable feedback-based control,} and (b) obstacle detection to avoid potential collisions. Second, we demonstrate a working system that can run all the RF processing on a embedded single-board computer (SBC) on the UAV and operate the UAV in real-time. In doing so, we identify and exploit trade-offs between sensing accuracy and computational overheads. The design of \name\ requires us to solve three challenges:

\para{Extracting Surface-Parallel Motion Insights: }\red{UAVs commonly use downward facing optical flow sensors (i.e. video cameras) to estimate ground-parallel motion. Optical flow sensors estimate the relative motion of the UAV by measuring the motion-induced visual shift between two consecutive image frames.} In \name, we replace this sensor with a downward facing mmWave radar which captures the reflections of the radar signal from the environmental reflectors (primarily the ground). Post-processing on these reflected signals reveals the range, angle, and doppler shift of each reflector. Recall that doppler shift is a frequency shift introduced by the motion of a reflector, and is therefore, a natural candidate for estimating UAV velocity. However, there is a key challenge. For a downward facing radar, the ground/floor moves parallel to the radar and any motion parallel to the radar does not introduce any doppler shift. So, how does one estimate the doppler shift for a surface-parallel motion?

Our first insight is -- at mmWave frequencies, \red{most} surfaces act as scatterers, not pure reflectors. Therefore, even seemingly flat surfaces, such as textureless floors have multiple weak reflectors. When the UAV moves, they exhibit unique patterns that reflect the direction of motion in the doppler-angle plane. We illustrate these patterns through an example in Fig.~\ref{fig:da_sim_vs_real}. The left figure plots a UAV flying towards right with velocity $v$. The doppler shift depends on the motion along the line joining each reflection point and the radar on the UAV. Therefore, a reflector at angle $\theta$ has a doppler shift $\frac{v\sin\theta}{\lambda}$, where $\lambda$ is the signal wavelength. This pattern is apparent in the center figure (built using simulation). This shows that the patterns in the doppler-angle plane correspond to the amplitude and direction of the UAV's velocity. %

\para{Multipath and Clutter: }The second challenge for designing radar-based perception stems from multipath and noise induced by objects in the environment. Objects in the environment introduce additional reflections, which appear to move in different directions and at varied velocities (similar to observing objects in a mirror). These reflections confound any RF-based perception system. This problem is escalated in cluttered indoor environments, which have many reflectors. This effect is demonstrated in Fig.~\ref{fig:da_sim_vs_real}(right) collected in a real-world setup. The figure exhibits the expected pattern but has other reflections and patterns embedded as well, largely due to clutter and multipath.

The effect of clutter and multipath makes it challenging to explicitly model the relationship between the doppler-angle plot and the velocity of the UAV. To extract insights from such cluttered data, we implicitly model this effect by using a data-driven approach. Specifically, we build a machine learning model that takes the doppler-angle heatmaps as inputs and outputs the estimated velocity of the UAV. We train this model using data from multiple environments and observe that it can learn an environment-invariant mechanism to translate the heatmaps to velocity, in spite of clutter. 

\para{Reducing System Overhead: }Finally, \name\ must be able run on small UAVs in real-time. \textred{The limited compute available to small UAVs introduces several practical constraints. For example, we cannot arbitrarily increase the size of doppler-angle heatmaps as it incurs too much data to process.} We observe that the design space of RF-based perception is quite large -- we can tune chirp parameters, frame rates, heatmap resolutions, model architectures, etc. Each parameter has consequences for the range of operation, sensing accuracy, and computational overhead. Our work explores this design space and identifies several optimizations to enable low compute overhead and \textred{can generate velocity estimates up to 50Hz}, sufficient to use existing motion control pipelines on UAVs. \textred{This ensures \name\ is reverse compatible with existing flight controllers, as we can simply replace optical sensor inputs with RF-based ones generated by \name.} 

\color{black}

We implement our end-to-end design on a small quadrotor built from commodity off-the-shelf parts. \red{We use two mmWave radars on this UAV as shown in Fig. \ref{fig:intro}}. We use a downward-facing radar to mimic a typical lidar altimeter and optical flow sensor used for \red{state estimates (velocity and altitude)}, and use a forward-facing radar to mimic a typical stereo camera used for collision detection. Our design can stably hover in place and fly given paths, while running all the compute on a low-power single-board computer on the UAV. Demos are available at the project website\footnotemark[1]. To summarize, our contributions are as follows.

\squishlist
    \item We identify and characterize the surface-parallel doppler-shift phenomenon at mmWave frequencies for a wide variety of surfaces, including those that are highly smooth, and show how this phenomenon can be seen in \red{doppler-angle heatmaps.}
    \item We train \red{a convolutional neural network (CNN)} to uncover the motion of the UAV with respect to the ground from these doppler-angle heatmaps, and show superior performance to state-of-the-art \red{commercial optical flow sensors} such as the PMW3901 in a wide variety of scenarios.
    \item \red{We explore various system design tradeoffs and optimizations needed to make \name\ compatible with existing flight controllers in a plug-and-play manner. We demonstrate the end-to-end system by deploying it on a real quadcopter. }
    \item We evaluate the use of radar as a collision avoidance sensor, and show \red{that it can be more effective than optical sensors in a wide range of realistic scenarios.}
\squishend

\section{Background}\label{sec:background}

We present a high-level background of key principles in optical flow and radar below.

\begin{figure}
    \centering
    \includegraphics[width=\linewidth]{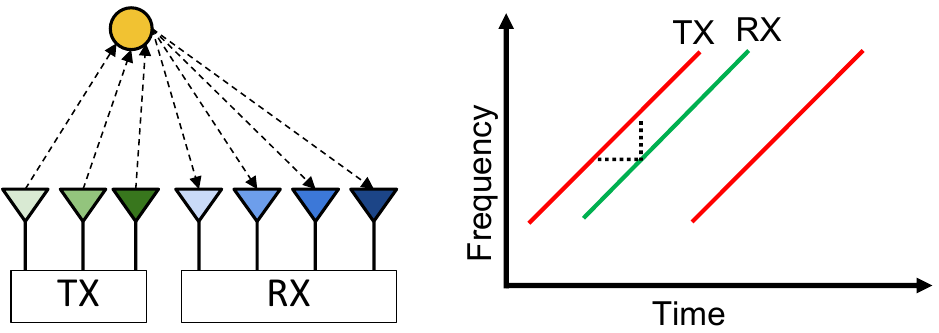}
    \caption{\red{\textit{Left.} A TDM MIMO scheme containing $3 \times 4$ TX-RX pairs. This allows for $N=12$ virtual antenna array for accurate angle estimation of a reflector (yellow). \textit{Right.} Sequence of transmitted FMCW chirps (red) and received echoes (green) for range and velocity estimation of reflectors. }}\vspace{-0.1in}
    \label{fig:tutorial}
\end{figure}

\subsection{mmWave Radar}
Recent years have seen a proliferation of single-chip mmWave radar sensors~\cite{felic2015single, lien2016soli}. A typical sensor leverages Frequency Modulated Carrier Wave (FMCW) chirps, wherein the frequency of the signal varies linearly with time \red{(Fig.~\ref{fig:tutorial} right)}. The sensor emits a sequence of FMCW chirps \red{into an environment} and captures the reflected signal at a set of receive antennas. \red{The distance of the reflectors can be inferred from the frequency shift between the transmitted and received chirps, and their bearing angle and relative velocity can be discerned from the apparent phase changes across consecutive receive antennas and chirps respectively. For brevity, we refer the reader to \cite{ti_tutorial} for a thorough treatment of FMCW radar signal processing fundamentals. For the rest of the paper, we will assume the convention that the unit of radar data for post-processing (i.e. radar cube) consists of a 3-D matrix of complex samples of dimension $N_c \times N_a \times N_s$, where $N_c$ is the number of chirps, $N_a$ is the number of receive antennas, and $N_s$ is the number of time samples taken per chirp interval at each receive antenna.}

\red{As explained in \cite{ti_tutorial}, the angular resolution of a linear antenna array for a fixed inter-antenna spacing improves with the number of antennas, $N_a$. Most inexpensive single-chip radars have only a small number of physical receive antenna in any given direction (i.e. 2-4 in Fig~\ref{fig:antenna} left), leading to poor angular resolution. A commonly used technique to improve angular resolution is to use time-division-multiplexing multiple-input multiple-output (TDM MIMO). Under TDM MIMO, each transmit antenna takes turns transmitting chirps into the environment. Hence, each distinct transmit-receive pair can be considered a separate (virtual) receive antenna (Fig.~\ref{fig:tutorial} left). This enables one to extend the number of antennas in a given direction, or to create a 2-D antenna array for simultaneous azimuth and elevation angle estimation (Fig~\ref{fig:antenna} right).}

\subsection{\red{Optical Flow Sensors}}

\begin{figure}
    \centering
    \includegraphics[width=\linewidth]{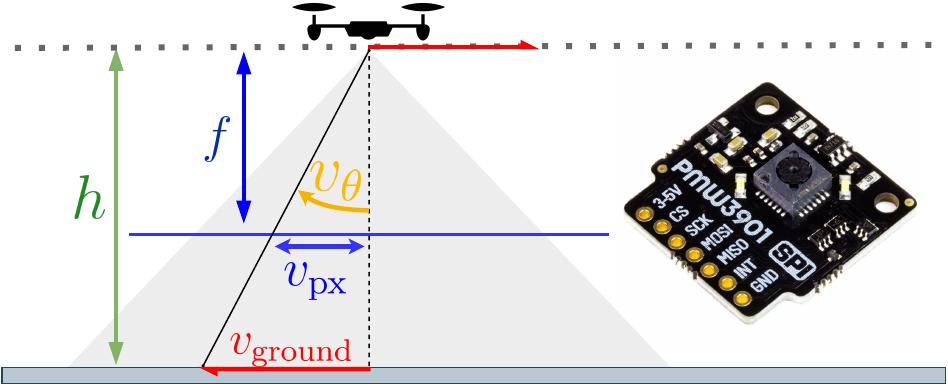}
    \caption{\red{An optical flow sensor first converts averaged optical flow (units in px) to an angular value $v_{\theta}$ using a known focal length $f$. This is combined with altitude $h$ to determine ground velocity.}}\vspace{-0.1in}
    \label{fig:optical_flow}
\end{figure}

\red{Optical flow refers to the apparent motion of visual features between two consecutive images.} Optical flow sensors estimate the shift in camera view over time and use it to estimate the motion pattern of the UAV. For example, if the UAV moves right, it sees the view of its optical flow sensor moving left. This feedback is provided to the flight controller on the UAV and is useful for two functions: (a) hovering -- the ability of the UAV to hover in-place when not required to move; and (b) motion feedback -- e.g., if the drone is trying to move forward, is it really moving forward? \red{The operation of such sensors on a UAV is depicted in Fig.~\ref{fig:optical_flow}. Algorithms for estimating optical flow have been studied in computer vision for decades \cite{lucas_iterative_1981, horn_determining_1981}. It is now common to find commercial optical flow sensors that integrate a small camera and an optical flow ASIC in a single package. It remains the most common and inexpensive option to achieve self-stabilization on small UAVs in GPS-denied environments. However, optical flow sensors fail in the dark, or when there is a lack of trackable visual features (e.g. smooth concrete floor). These failure modes are shared by more complex and expensive classes of camera-based techniques for estimating egomotion in GPS-denied environments, i.e. visual odometry (VO) and visual SLAM (VSLAM).}

\section{Motivation and Goals}

\red{Our high level goal is to enable collision-free flight through general unknown environments that may contain static or dynamic obstacles using only onboard radar perception. We envision that this would enable a plethora of use cases that conventional optical sensor + GPS reliant systems struggle with or find impossible. For example, being able to fly in foggy conditions at dawn, navigate in urban environments without crashing into the glass exteriors of modern buildings or power lines, and enabling drone deliveries over residences while respecting privacy. Our strategy involves two steps:}

\squishenum
\item \red{We first achieve simultaneous altimetry and horizontal velocity estimation using a single downward facing radar. This is key since unlike ground vehicles, small quadcopters are innately unstable and will constantly drift unless given motion feedback from the sensing the environment. Without being able to measure its own velocity and altitude, a quadcopter instructed to merely loiter in place will drift off within seconds. Similarly, the quadcopter will be unable to follow more complex commands without motion feedback. }
\item \red{Once self-stabilization is achieved and velocity-based control becomes possible, we fix the drone's altitude and use a forward facing radar to sense the environment and navigate the drone in the 2-D plane corresponding to this fixed altitude. This simplifying assumption is sufficient to test the capabilities and limitations of the system.}
\enumend

\subsection{Design Choices}
\noindent\textbf{Why not use the front radar for odometry? } Drone primarily rely on their downward-facing optical flow sensor at low altitudes where GPS is unavailable. At low altitudes, drones can encounter dynamic obstacles (e.g. moving humans, cars) in front of or besides the drone. These dynamic objects interfere with visual odometry pipelines which rely on a static world assumption \cite{minoda_viode_2021}. Such dynamic objects will likely introduce errors in radar perception as well. Conversely, a forward facing radar might not encounter any reflectors at all in the environment (e.g. in an open field), leading to a lack of features to perform odometry. On the other hand, at low altitudes a strong ground reflector is always available. Furthermore, even if a forward facing radar is used to obtain odometry estimates, an auxiliary downward facing sensor is still necessary for altimetry. Hence, by using a single downward facing radar to estimate both altitude and horizontal velocity estimates, we prevent any redundancy.

\para{Why not rely on other tracking solutions (e.g. UWB)? } Such schemes assume that there is pre-existing infrastructure deployed in the environment, such as UWB anchors and IR cameras. This is infeasible for the vast majority of real world environments that one would encounter in the wild. As such, we do not consider infrastructure-assisted approaches to be fully autonomous.

\para{Is velocity-based control sufficient for navigation? } There are several ways of navigating a drone. In classical approaches \cite{hart_formal_1968, LaValle1998RapidlyexploringRT}, there is a preexisting map and a desired end point is specified on this map. A path planning algorithm outputs a trajectory (i.e. sequence of coordinates on the map) to the end point. The drone constantly localizes itself on the map using some technique (e.g. SLAM, GPS). The positional error to the next waypoint is repeatedly measured and minimized using a low-level controller i.e. proportional–integral–derivative (PID) control~\cite{johnson2005pid}. This paradigm is infeasible for our case since we assume our environment is unknown (we do not have a preexisting map). Indeed, it would be impossible to extensively map out all the environments our drone could possibly fly in. 

However, there exists a map-free paradigm for navigation in the form of velocity-based control, which has recently become popular \cite{gandhi_learning_2017, sadeghi_cad2rl_2017, bonatti_learning_2020, kaufmann_deep_2018} . In this paradigm, commands are relative to the drone's current position (e.g. move forward) rather than being based on an absolute global coordinate system. Hence, localization in any global reference frame is not necessary. However, the drone needs velocity-feedback to ascertain that it is following the commands correctly. To navigate, all the drone must do is to constantly react to its sensory input (i.e., front radar) in a sensible manner. This is sufficient for certain key tasks such as lane or tunnel following.  \name\ provides velocity-based feedback to the drone, but is agnostic to how these commands are generated.

\section{Surface-Parallel Doppler Shift}\label{sec:doppler}

\begin{figure}
    \centering
    \includegraphics[align=c, width=\linewidth]{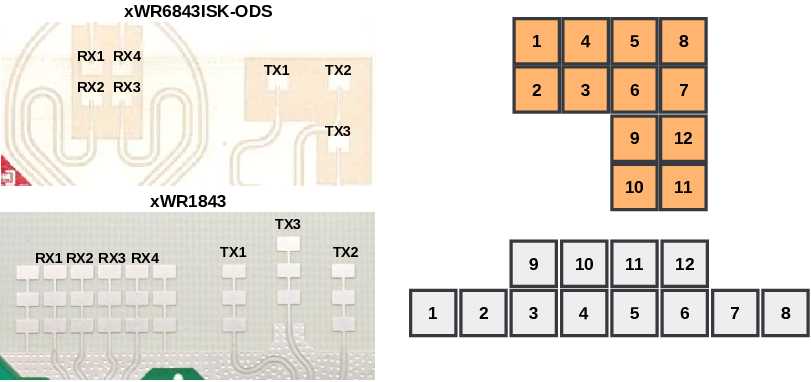}
    \caption{\red{\textit{Left.} Physical antenna array layouts on single-chip mmWave radar boards. \textit{Right.} Corresponding numbered virtual antenna array under TDM MIMO.}}\vspace{-0.1in}
    \label{fig:antenna}
\end{figure}

Our first goal is to estimate the relative motion of the UAV with respect to its environment \red{using a downward facing radar. A naive solution is to directly adapt a scan matching approach based on point clouds as is done with lidar. This would involve detecting individual reflectors and tracking their relative motion across a sequence of measurements. However, mmWave point clouds are extremely sparse and unstable compared to lidar point clouds, making scan matching extremely difficult \cite{lu_milliego_2020}. This issue is further exacerbated when pointing radar to a geometrically featureless flat surface (i.e. the ground).} However, FMCW radar offers us a better option: doppler shift. Doppler shift directly captures the velocity of reflectors with respect to the UAV. Intuitively, if the radar is moving away from a reflector, the doppler shift is negative. If the radar is moving towards the reflector, the doppler shift is positive.

\para{\red{Seeing Surface Parallel Motion:}} The key challenge for our design is that the motion of the UAV is parallel to the \red{ground}. Specifically, for a reflector placed at angle $\theta$ with respect to the radar (see Fig.~\ref{fig:da_sim_vs_real}), the doppler shift \red{corresponds to a velocity of $v \sin\theta$}, where $v$ is the \red{radial velocity} of the object. Therefore, when a radar is facing downwards, the primary reflector, i.e. the floor, is at angle $\theta=0^{\circ}$ and induces a zero doppler shift. 

\red{Our first insight is that due to the high frequency of mmWave signals, large reflectors like floors induce scattering, as shown in Fig.~\ref{fig:da_sim_vs_real}. Specifically, they can feature many weak reflectors at oblique angles with respect to the radar. If we densely aggregate the doppler shift for these reflectors in a \red{doppler-angle} heatmap, it should appear as a continuous curve that varies as $v \sin\theta$, where $v$ is the velocity of the drone.}

\para{\red{Dealing with Low Resolution: }} \red{But what does this pattern look like under realistic hardware constraints? Note that the antenna arrays on single-chip radars (illustrated in Fig.~\ref{fig:antenna}) usually falls under one of two categories: (a) having a 1D antenna array with 8 virtual antennas in the azimuth direction~\cite{iwr1843boost}, or (b) having a 2D antenna array but only having 4 virtual antennas in the azimuth and elevation direction~\cite{iwr6843isk-ods}. We need to choose option (b) since we must estimate motion along two degrees of freedom in the ground-parallel plane. As explained in Sec.~\ref{sec:background}, this leads to an extremely low angular resolution of nearly $1/2$ radians ($30^\circ$) at best.}

\red{Our second insight is that although resolution in the angle axis is indeed limited, we can compensate for this by arbitrarily increasing the doppler axis. We demonstrate this effect by simulating a length 4 uniform antenna array in motion in Fig.~\ref{fig:da_sim_vs_real} (center). As shown, the boundaries of the pattern become well resolved with a sufficiently large doppler resolution. Specifically, we find that the expected doppler-angle heatmap takes the form of a fat sinusoidal shape with smooth edges. This shows that the doppler-angle heatmap representation is both evocative and informative in spite of low angular resolution.}

\color{black}
\section{Learning-based Flow Estimation and Collision Avoidance}
As mentioned in Sec.~\ref{sec:intro}, real-world measurements of radar signals are prone to multipath effects and clutter. Intuitively, multipath creates reflections in the environment that appear to be at different locations and move at different velocities than real objects. This is analogous to looking at a mirror. If you observe an object approaching the mirror, the object and the reflection appear to be moving in opposite directions. Similarly, clutter leads to overlap of signal from multiple reflectors in the environment at the radar, making it harder to disentangle individual patterns. 
Instead of modelling these effects explicitly, we learn an implicit model of these effects by leveraging a data-driven approach. This is consistent with recent work in this space. Specifically, ~\cite{bansal_pointillism_2020,lu_milliego_2020,lu_see_2020} use a data-driven learning-based approach for different radar sensing applications. 

Our approach differs in the following way: such past work uses a sparse point cloud input representation, where each point corresponds to a potential reflector in the environment. As this format mimics lidar-like outputs but is orders of magnitudes more sparse, applying existing lidar point cloud processing techniques to such data remains exceedingly challenging. On the other hand, our work leverages dense doppler-angle heatmaps. Since such heatmaps are images, they are directly compatible with mature and highly optimizable convolutional architectures commonly used in computer vision. We opt for an all heatmap and CNN-based approach, leveraging doppler-angle heatmaps and range-angle heatmaps for velocity/flow estimation and collision detection respectively. We show that this heatmap-based approach is a uniquely advantageous, as unlike past work, we can deploy our models on a small UAV and operate at high update rates (up to 50 Hz).
\color{black}

\subsection{Radar Selection}
As noted before, \name\ uses two mmWave radars -- one forward facing and one downward facing. There are multiple off-the-shelf radars available with different antenna layouts, frequency bands, etc. We choose TI xWR1843 \cite{iwr1843boost} for the forward facing radar. This is because this radar has a large linear virtual antenna array. It has an antenna array comprised of 3 Tx antennas and 4 Rx antennas, i.e., \red{8} effective antennas arranged horizontally in a line. The large horizontal aperture provides very fine-grained azimuth resolution. This is essential for the UAV as it needs to move left or right depending on potential for collision along a given direction. The high angular resolution for this radar allows the UAV to make subtle shifts in its heading direction and effectively avoid collisions.

On the other hand, the downward facing radar needs to identify UAV motion parallel to the ground for motion flow estimation. This motion can happen in any direction parallel to the ground, which means we require a 2-D antenna array. This functionality is provided by the TI xWR6843ISK-ODS \cite{iwr6843isk-ods}. The antenna pattern is shown in Fig.~\ref{fig:antenna}. Since each antenna array is much shorter, we sacrifice angular resolution for the ability to measure both azimuth and elevation angle with respect to the radar simultaneously.

We run independent compute pipelines on each of these radars on a low-end compute board present on the UAV. This board feeds the outputs to an unmodified flight controller. 

\subsection{\red{Estimating Radio Flow}}

\begin{figure*}[htp]
    \centering
    \includegraphics[align=c, width=0.9\linewidth]{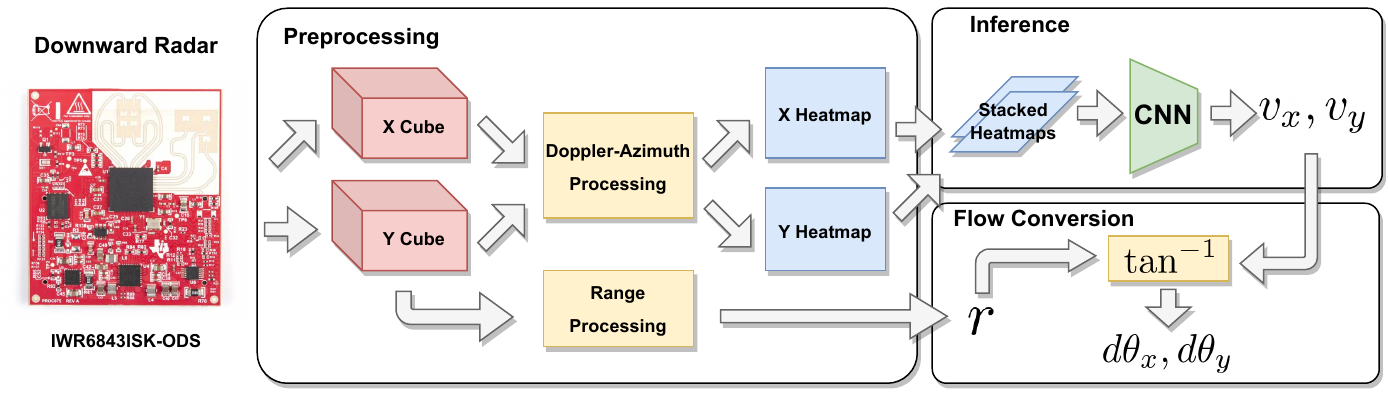}\vspace{-0.1in}
    \caption{\red{\textbf{Flow prediction pipeline.} \textit{Left.} Preprocessing stage where an incoming radar cube is turned into stacked doppler-angle heatmaps and the altitude above ground is estimated. \textit{Top right.} Velocity prediction using a neural network. \textit{Bottom right.} Converting velocity predictions to angular flow rate using the altitude estimate and trigonometry.}}
    \label{fig:flow_model}\vspace{0.1in}
\end{figure*}

First, we describe our \red{radio flow} estimation pipeline (outlined in Fig.~\ref{fig:flow_model}). The flow estimation pipeline consists of three stages: (a) a preprocessing stage, where signal processing functions like doppler-angle heatmaps and altitude estimates are extracted from the radar samples, (b) an inference stage, where the heatmaps are given to a learned model to infer the current ground velocity, and (c) where the outputs of the inference and the altitude estimates are combined using trigonometry to obtain the final angular flow values. We note that, in principle, the UAV just needs to estimate its velocity. However, we perform the third step in order to integrate our system into an unmodified flight controller, which is tuned to receive optical flow values and range estimates independently.

\SetKwFunction{SubsampleRange}{SubsampleRange}
\SetKwFunction{DopplerAngleFFT}{DopplerAngleFFT}
\SetKwFunction{SumOverRange}{SumOverRange}
\SetKwFunction{Normalize}{Normalize}
\SetKwFunction{Stack}{Stack}
\SetKwFunction{Resize}{Resize}
\begin{algorithm}[t]
\KwIn{Radar sample cubes $\mathbf{S}_x, \mathbf{S}_y \in \mathbb{C}^{N_{c} \times 4 \times N_{s}}$}
\Parameter{Angular range $\theta$ and resolution $d\theta$, range subsampling factor $K$, resize shape $M,N$}
\KwOut{Stacked heatmap $\mathbf{H} \in \mathbb{R}^{H \times W \times 2}$}

$\mathbf{S}_x \gets$\SubsampleRange$(\mathbf{S}_x,K)$\;
$\mathbf{S}_y \gets$\SubsampleRange$(\mathbf{S}_y, K)$\;

$\mathbf{S'}_x \gets$\DopplerAngleFFT$(\mathbf{S}_x, d\theta, \theta)$\;
$\mathbf{S'}_y \gets$\DopplerAngleFFT$(\mathbf{S}_y, d\theta, \theta)$\;

$\mathbf{H}_x, \mathbf{H}_y \gets$\SumOverRange$(\mathbf{S'}_x)$, \SumOverRange$(\mathbf{S'}_y)$\;
$\mathbf{H}_x, \mathbf{H}_y \gets$\Normalize$(\mathbf{H}_x)$, \Normalize$(\mathbf{H}_y)$\;

$\mathbf{H} \gets$\Stack$(\mathbf{H}_x, \mathbf{H}_y)$\;

$\mathbf{H} \gets$\Resize$(\mathbf{H}, M, N)$\;

\Return $\mathbf{H}$
\caption{Doppler-Angle Preprocessing}
\label{alg:preprocessing}
\end{algorithm}

\para{Preprocessing:} Our preprocessing stage receives a \red{3-D matrix (i.e. radar cube) $\mathbf{S} \in \mathbb{C}^{N_c \times N_a \times N_s}$ of samples from the radar, where $N_c$ is the number of chirps, $N_a$ is the number of (virtual) receive antennas, and $N_s$ is the number of time samples taken per chirp interval. Since we are using TDM MIMO on the xWR6843ISK-ODS, $N_a = 12$ as shown in Fig.~\ref{fig:antenna}. On the other hand, $N_c$ and $N_s$ are user configurable parameters. Increasing $N_c$ and $N_s$ increases accuracy at the expense of computational overhead (Sec.~\ref{sec:design} and \ref{sec:results}). }

\red{We start by computing altitude from $\mathbf{S}$. Then we split $\mathbf{S}$ into orthogonal 1D X and Y ground-parallel components. We do this by choosing  and combining samples from specific numbered RX antennas (Fig.~\ref{fig:antenna}).}Specifically, we mimic a 1D antenna array in the X direction by stacking and adding the samples received at antennas 1, 4, 5, 8 and 2, 3, 6, 7 in those specific orders. We do the same thing for the Y direction using antennas 8, 7, 12, 11 and 5, 6, 9, 10. \red{This gives us $\mathbf{S}_x, \mathbf{S}_y \in \mathbb{C}^{N_c \times 4 \times N_s}$. }

We feed \red{$\mathbf{S}_x, \mathbf{S}_y$} into the remainder of the preprocessing stage \red{(Algorithm \ref{alg:preprocessing})}. \red{First, the $\mathbf{S}_x$ and $\mathbf{S}_y$ radar cubes are downsampled along the range axis by a tunable factor $K$. By reducing the amount of samples, we reduce the computational burden of the following step. Next, we compute doppler-angle heatmaps at each of the $N_s / K$ range bins of $\mathbf{S}_x, \mathbf{S}_y$ using a 2-D FFT. After this, summing over the range bins gives us doppler-angle heatmaps. Then we normalize the heatmaps such that all pixels lie within $[0.0, 1.0]$. Finally, the normalized X and Y heatmaps are stacked into a 2 channel image $\mathbf{H}$, which is then downsampled to a smaller size via interpolation. $\mathbf{H}$ is then fed to a feedforward model for inference.}

\para{Inference:} Our goal is to extract 2-D velocity information from the stacked heatmaps $\mathbf{H}$, i.e. velocity along X and Y axes. From Sec.~\ref{sec:doppler}, we know that the velocity information exhibits as the amplitude of a sinusoidal curve in these maps. To extract this information, we opt to use a \red{CNN}.

For the model architecture, we consider several different architectures based on paring down ResNet18 in order to control the computational cost of an inference. We choose ResNet18 as the base model as it is exhibits good accuracy while still respecting the computational constraints of SBCs on small UAVs. We call the pared-down models ResNet18Mini (4 layers), ResNet18Micro (2 layers), and ResNet18Nano (1 layer). Aside from removing layers, \red{our previous choice of downsampling $\mathbf{H}$ also plays a vital role in reducing the computational cost. }

\para{\red{Angular Flow Conversion:}} A standard flight controller takes in optical flow values, which are measured as angular values (since absolute height information is unavailable to an optical sensor alone). We plan to convert the velocity measurements into these angular values for compatibility with off-the-shelf controllers. One might wonder if it is possible for the model to regress to the angular flow values directly. However, this is not possible because the doppler-angle heatmaps correspond to absolute velocities. We include a angular flow conversion step which normalizes the output of the model into angular flow values using the current estimated altitude above ground (in preprocessing) and a basic trigonometry identity. %

\begin{figure}
    \centering
    \includegraphics[width=.46\linewidth]{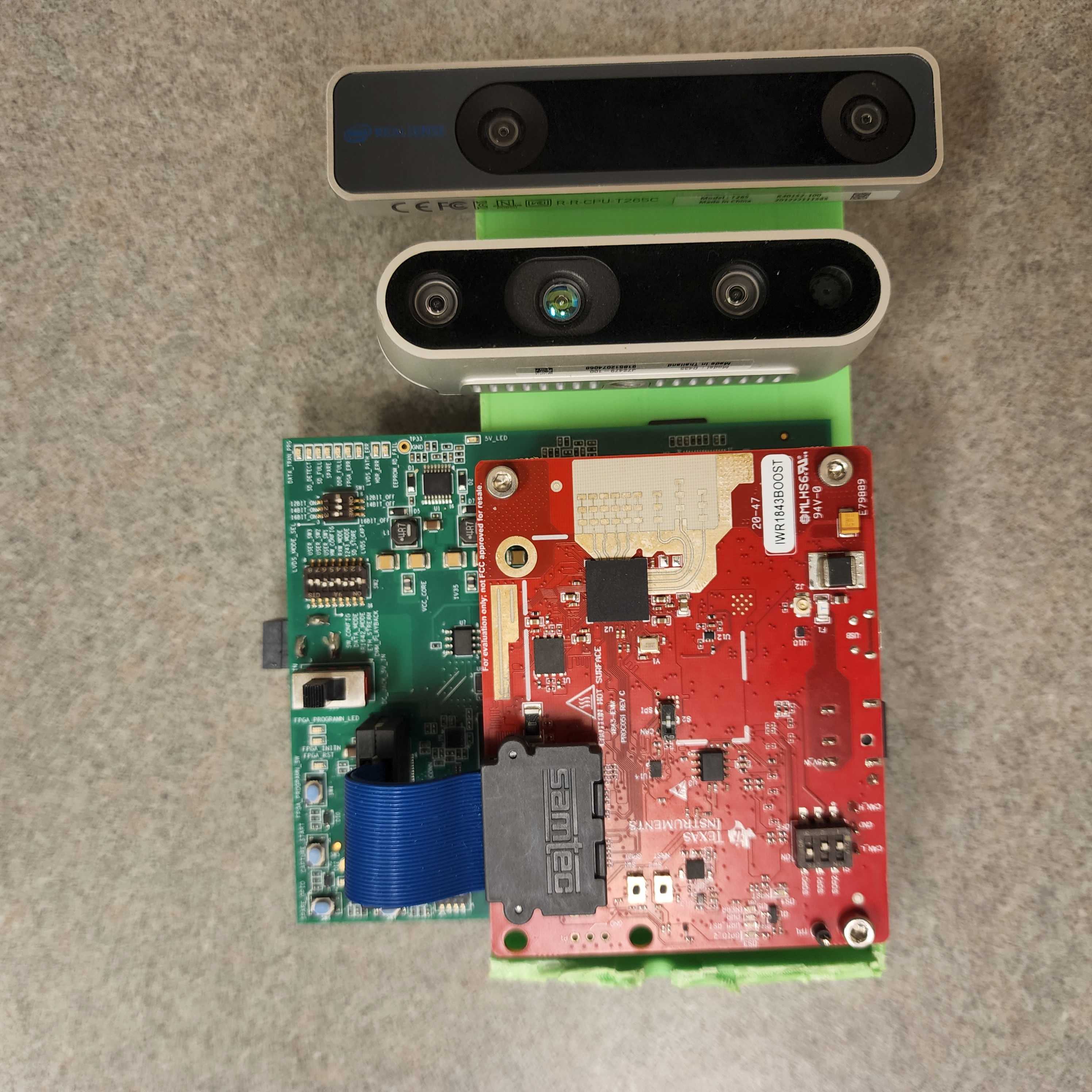}
    \includegraphics[width=.46\linewidth]{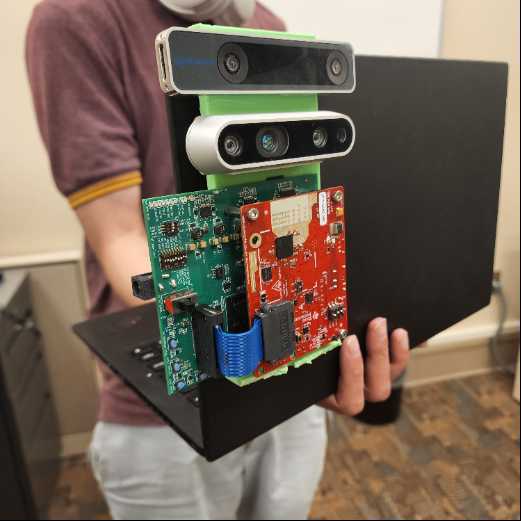}
    \caption{\textbf{Data collection apparatus.} \textit{Left:} We attach a RealSense T265 tracking camera, a D435 stereo depth camera, and a DCA1000 capture card to a 3D-printed frame. We use both an IWR1843 and an IWR6843ISK-ODS (not shown) with the DCA1000. \textit{Right:} Moving the sensor suite around by hand. }
    \label{fig:data_collection}
\end{figure}

\para{Dataset Collection:} We aim to make the training data collection process seamless and low overhead. Therefore, we design the sensor suite depicted in Figure \ref{fig:data_collection}, which consists of an Intel RealSense T265 \cite{t265} tracking camera, a D435 \cite{d435} stereo depth camera, and a DCA1000 data capture card \cite{dca1000} attached to a 3D-printed frame. We configure the T265, D435, and DCA1000 to stream 6DoF pose readings, images, and radar cubes at a rate of 200 Hz, 30 Hz, and up to 50 Hz respectively. 

In order to collect a dataset for our model, we point the sensor setup to random admissible flat surfaces in a wide variety of indoor and outdoor environments. Once a surface is chosen, we record a sequence of timestamped poses, depth images, and radar samples while moving the sensor for a fixed length of time (typically 100 seconds). The motion undertaken can be arbitrary as long as the sensor suite is continuously pointed towards the surface, however the goal is to cover as wide of a range of motions in the plane parallel to the surface as possible and to mimic the range of motion that a drone might take during flight. Then, we compute the ground truth angular flow at each time using the pose estimates from the tracking camera as well as the range readings from the depth camera, and use nearest-neighbor interpolation to assign a ground-truth for each radar cube. We ensure that our dataset consists of traces of a wide variety of surfaces, such as carpets, grass, concrete, drywall, metal, and so on. To train each model, we collect at least 25 traces of surfaces of 100 seconds each. For a radar capturing at a rate of 50Hz, this translates to 125000 datapoints in total per dataset.

\para{Model Training:} To train a model, we first split our dataset into training and validation sets of surfaces in a 60/40 ratio. For our loss function, we use the RMSE (L2) loss function, which is common for regression tasks, and use Adam optimizer~\cite{kingma2014adam}. We find that using a batch size of $128$ and a learning rate of $1\text{e}-4$ works well across all model architectures. At the end of each epoch, we log the RMSE of the model's prediction on the validation set, and keep the model with the best performance. We typically do not need to train for any more than $200$ epochs to saturate the performance on the validation set. The above is done for one set of preprocessing parameters and choice of model architecture - to find the best set of preprocessing parameters and choice of model architecture, we run a grid search on a range of possible parameter choices and choose the model with the best validation performance.

\subsection{Detecting and Avoiding Collisions}

\begin{figure}
    \centering
    \vspace{-0.1in}
    \includegraphics[width=\linewidth]{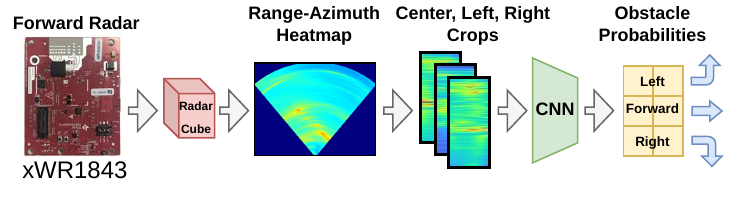}\vspace{-0.1in}
    \caption{\red{\textbf{Collision detection.} We use a CNN that classifies a the probability of an obstacle in the left, center, and right part of a color image, depth image, or radar heatmap.}}\vspace{-0.1in}
    \label{fig:collision_avoidance}
\end{figure}

With the velocity estimation from the downward facing radar, we can instruct the UAV to follow velocity setpoints. For example, we can have the UAV move forward at a constant speed, stop, turn, and move in another direction. But how do we use the forward facing radar to instruct the UAV to avoid and navigate around obstacles in the environment?

\para{Intuition: }Intuitively, collision detection relies on finding obstacles in close proximity. For collision avoidance task, we compress our radar data into a range-angle plot. At a high level, our goal is to identify if there is an obstacle directly in front of the radar, i.e. at a short range. If a possible collision is detected, we want to identify the direction to steer to (i.e., what direction is collision-free?). We note that we expect radar sensors to fare better than visual sensors in scenarios like glass windows or mirrors, which fail to register as obstacles for visual sensors.

\para{Design:} Our learning approach is inspired from \cite{gandhi_learning_2017}, which trains a deep neural network that takes a color image from the forward facing camera, crops the image into a center, left, and right component, and outputs the probability of there being an obstacle in a small distance ($\leq\sim$ 1m) in the direction of each component. The idea is to move in the direction with the lowest probability of containing an obstacle. For example, if the center crop has an obstacle probability below a certain threshold, the UAV moves forwards. However, if it is blocked, it stops and turns in the direction of the crop with the lowest obstacle probability.

We adapt this basic design by modifying the input to take in 2D range-angle heatmaps from the forward facing radar instead of RGB images from a forward facing camera. We first derive the range-angle heatmaps from the radar samples using Capon beamforming~\cite{capon1969high} for an angle range of $\pm 45$ degrees from boresight and an angular resolution of $1$ degree. Then, we split the resulting heatmap into a left, center, and right crop each spanning $30$ degrees. Finally, the crops are given to a a neural network as a single batch to predict the probability of an obstacle in each crop.

\para{Model Architecture:} Our model architecture is again based on ResNet18~\cite{he2016deep}. Specifically, we modify the first layer to take into account that our range-azimuth heatmaps are single-channel images, and modify the final fully connected layer to output two logits representing the probability of there being an obstacle and not being an obstacle respectively. We train the network using the standard cross-entropy loss function for classification tasks.

\para{Dataset Collection:} To collect the dataset, we use the same sensor setup as depicted in Figure \ref{fig:data_collection}. We move the sensors around by hand in a large variety of real world environments and conditions. But how do we label the ground-truth for each crop of each range-azimuth heatmap? To find range-azimuth crops containing obstacles, we collect trajectories where the radar is pointed at random obstacles (i.e. going down the side of a building, towards the wall of a hallway while walking through it) and label the crops of all collected range-azimuth heatmaps as containing obstacles. Conversely, to find obstacle-free range-azimuth heatmap crops, we collect trajectories keeping the radar far away from any obstacles (i.e. wandering in an open atrium). To prevent bias in the dataset, we collect an equal number of obstacle-filled and obstacle-free range-azimuth heatmap crops.

\para{Model Training:} Similar to the flow prediction model, we split our dataset into a training and validation set of image crops in a ratio of 80/20. We train the models using the standard cross-entropy loss function for classification tasks, with Adam as the optimizer, a learning rate of $1\text{e}-4$, and a batch size of $128$. To prevent overfitting, we do not train the model for more than 20 epochs, and keep the model with the best performance on the validation set.

\section{System Design}\label{sec:design}

We discuss our experimental UAV platform and the design
decisions involved.

\subsection{\red{Assembling the Quadcopter}}

\begin{figure}
    \centering
    \includegraphics[width=.45\linewidth]{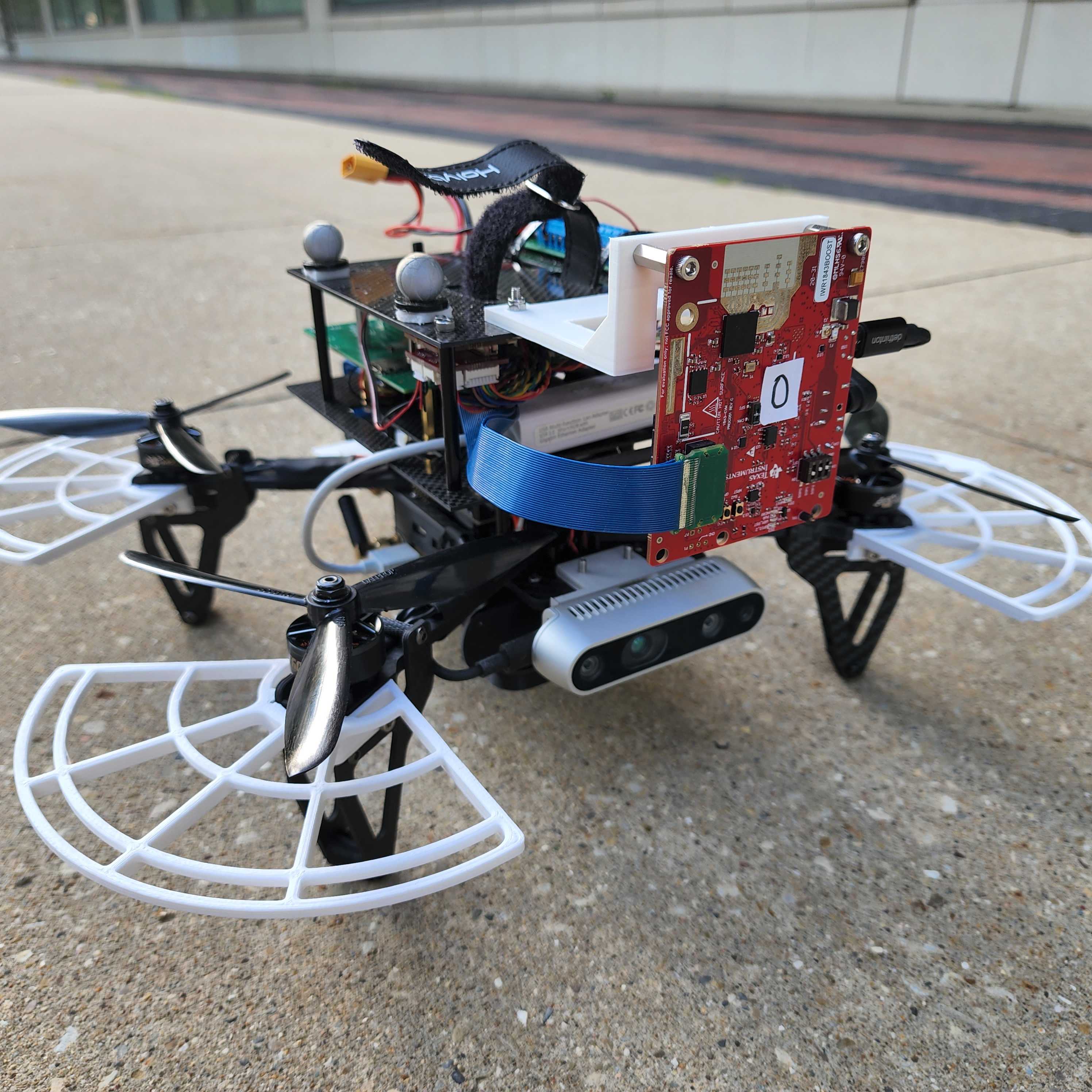}
    \includegraphics[width=.45\linewidth]{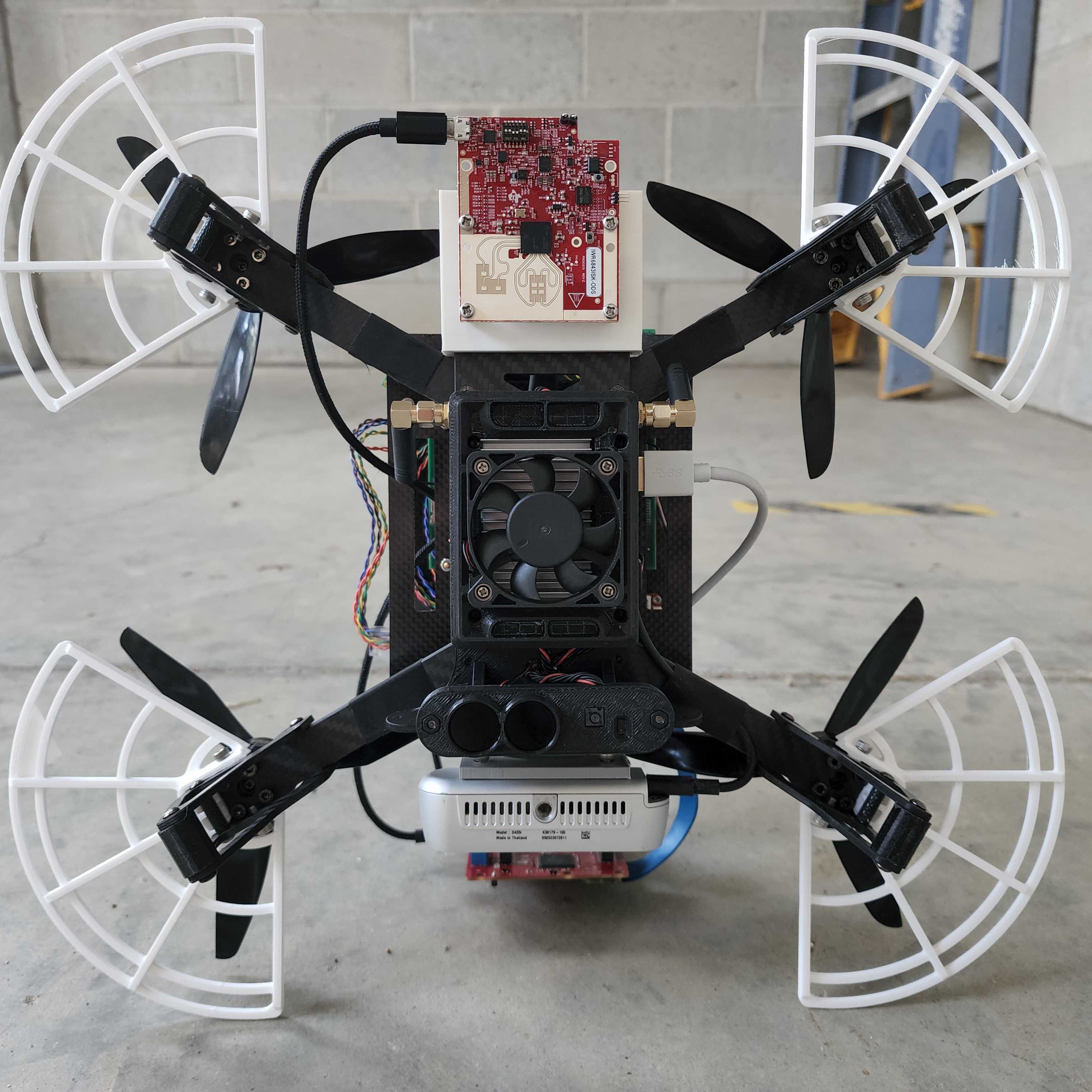}\vspace{-0.1in}
    \caption{\textbf{Our experimental UAV platform.} \textit{Left:} Frontal view, featuring \red{the IWR1843 and the D435i camera}. \textit{Right:} Bottom view, featuring the optical flow and rangefinder module \red{(bottom)} and the AWR6843ISK \red{(top).}}\vspace{-0.1in}
    \label{fig:px4}
\end{figure}

We assembled our experimental UAV platform shown in Fig.~\ref{fig:px4} from commercial off-the-shelf parts. We use a frame with a relatively small wingspan of 300mm, which allows us to fly in constrained environments. For the flight controller, we use the Pixhawk 4 Mini \cite{pixhawk4}, which runs the open source PX4 Autopilot flight stack firmware \cite{px4autopilot}.

\para{Sensors:} For comparison, the UAV features two independent sensor suites, a typical optical sensor suite and our own radar-based sensor suite. Only one sensor suite is used at a time throughout our experiments. 

The baseline optical sensor suite consists of a downward-facing PSK‐CM8JL65‐CC5 unidirectional infrared rangefinder \cite{cm8jl65} and PMW3901 optical flow sensor \cite{pmw3901} for state-estimation and a forward-facing Intel RealSense D435i~\cite{d435} stereo depth camera for collision avoidance.

The radar sensor suite consists of a downward-facing 60GHz TI AWR6843ISK single-chip radar \cite{iwr6843isk-ods} and a forward-facing 77GHz TI IWR1843 single chip radar \cite{iwr1843boost} for state-estimation and collision avoidance respectively. We pair each radar with a DCA1000EVM \cite{dca1000} board in order to extract raw radar samples rather than preprocessed point clouds from the radars.

\para{Onboard Compute:} For compute, we use an Up Board \cite{upboard} single-board computer (SBC) running Ubuntu 18.04, featuring a quad-core Intel Atom x5-z8350 CPU and 4 GB of RAM. The SBC is connected to the two DCA1000 boards via a 1Gb Ethernet interface, which allows for continuous streaming of radar samples from the boards without dropped packets. It is also connected to the Pixhawk 4 flight controller through a UART interface. We use this to send the flight controller angular flow estimates from our radar flow pipeline as well as velocity setpoints from our collision avoidance module.

\subsection{\red{Flow Update Rate vs Accuracy}}

Recall that we use a sequence of \red{$N_c$ chirps} to compute the doppler-angle plots. \red{This, in addition to the chirp time $T_c$, determines the maximum unambiguous velocity $v_{\text{max}}$ and velocity resolution $v_{\text{res}}$.} Specifically,

\begin{equation}
    v_{\text{max}} = \frac{\lambda}{4T_c}, \quad v_{\text{res}} = \frac{\lambda}{2N_cT_c}
    \label{eq:doppler}
\end{equation}

\red{Note that $v_{\text{max}}$ and $v_{\text{res}}$ are inversely proportional to $T_c$ and $N_cT_c$ respectively.} In practice, \red{$T_c$} is limited by the capabilities of the radar frontend - for most single chip radars it is possible to choose \red{$v_{\text{max}}$} of anywhere between $1$ to $20$ m/s. \red{However, we} can arbitrarily increase the velocity resolution in principle \red{by simply increasing $N_c$ (i.e. accumulating more chirps). }

\red{We call the time needed to collect these chirps the integration time, i.e. $T_{\text{int}} = N_cT_c$. Furthermore, data arrives from the DCA1000 in batches of 20-30 chirps at a time, which is typically smaller than $N_c$. Whenever a new batch of chirp data arrives, we accumulate the chirp data into a FIFO circular buffer of size $N_c$ and then run the flow estimation pipeline using all the chirps accumulated in the buffer. Hence, the rate of arrival of new chirp data determines the flow update rate $f_{\text{flow}}$, and} the total latency of the system comprises \red{the integration time well as the flow estimation time. }

\red{Clearly, the flow estimation pipeline must complete before the next batch of chirp data arrives, otherwise some data will be dropped. Hence, higher update rates impose stricter computational requirements.} This leads to a tradeoff between \red{flow update rate and estimation accuracy}. The higher the value of \red{$N_c$}, the better the accuracy of our velocity estimates. However, it also increases \red{the system latency}. This means the reaction time of the UAV is increased. \red{By contrast, lowering $N_c$ allows for faster and lower latency updates at the expense of accuracy.}

We explore tradeoff empirically in Sec.~\ref{sec:tradeoff_empirical}. Specifically, we tune the radar processing pipeline and the ML models using the parameters in Tab.~\ref{tab:parameters} to achieve update rates of 15Hz to 50 Hz, and compare their performance.

\section{Evaluation}\label{sec:results}

We conduct our experiments in an $6 \times 6$ m indoor flight arena, which features a textureless soft foam floor interspersed with textured hard foam tiles which function as landing pads. We obtain the ground truth motion of the UAV throughout its flight using a VICON capture system. 

\begin{figure}
    \centering
    \includegraphics[width=.8\linewidth]{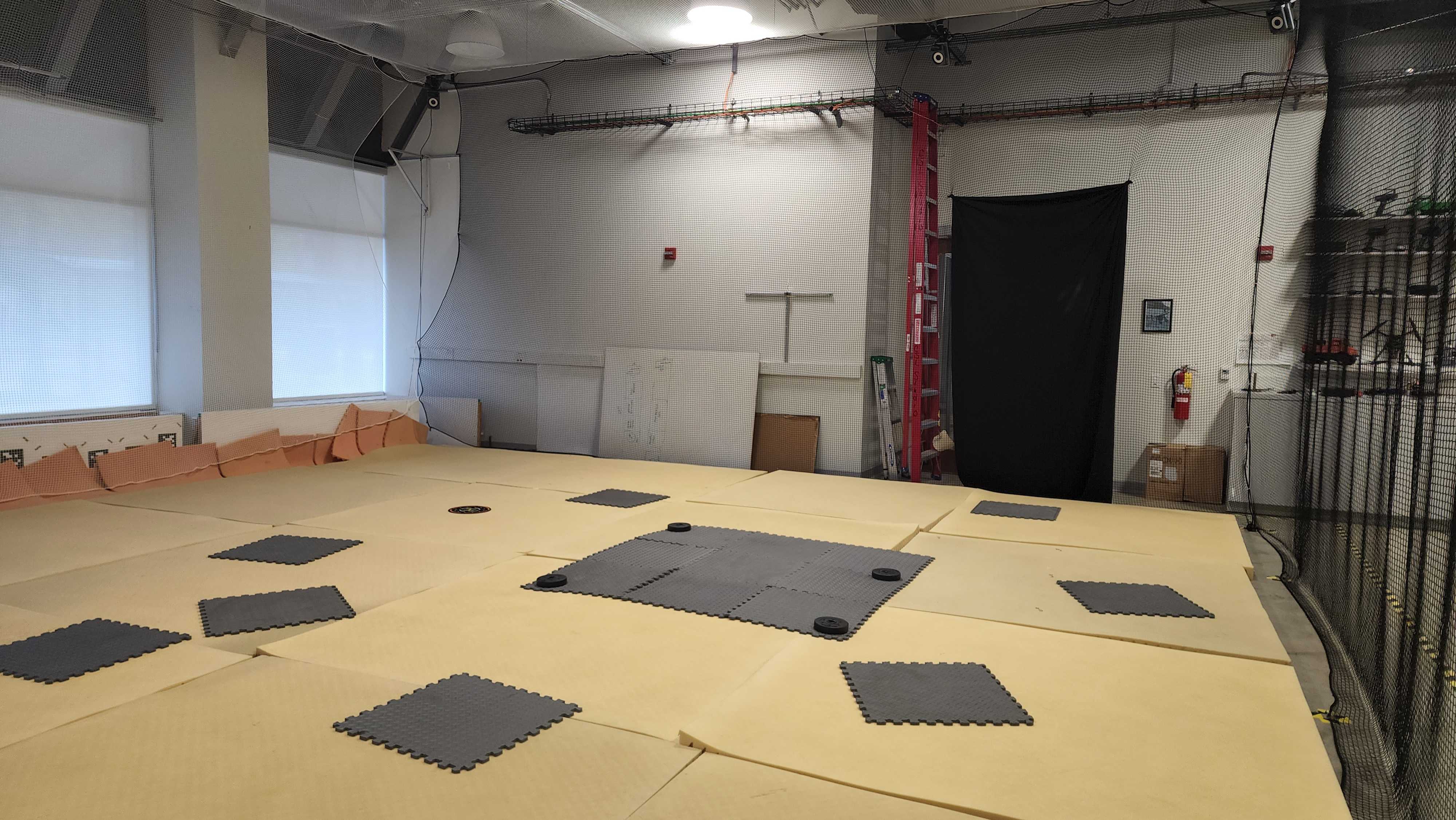}\vspace{-0.1in}
    \caption{\textbf{Indoor flight arena used for experiments.} VICON is used to measure the UAV's motion.}\vspace{-0.1in}
    \label{fig:px4}
\end{figure}

\begin{figure*}[htp]
\centering
\begin{subfigure}{0.30\textwidth}
\centering
\includegraphics[width=\textwidth]{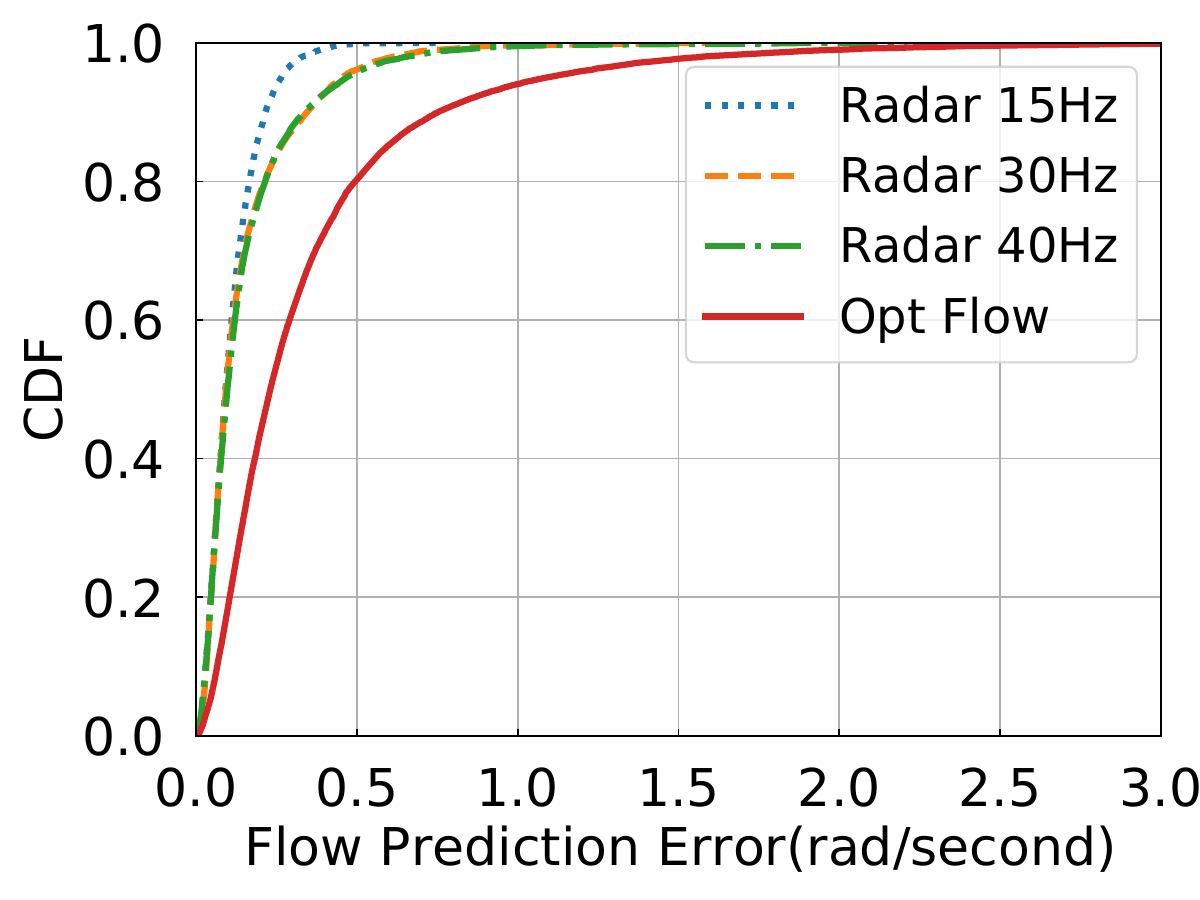}\vspace{-0.1in}
\caption{Flow prediction accuracy}
\label{fig:loiter_method}
\end{subfigure}
\quad
\begin{subfigure}{0.30\textwidth}
\centering
\includegraphics[width=\textwidth]{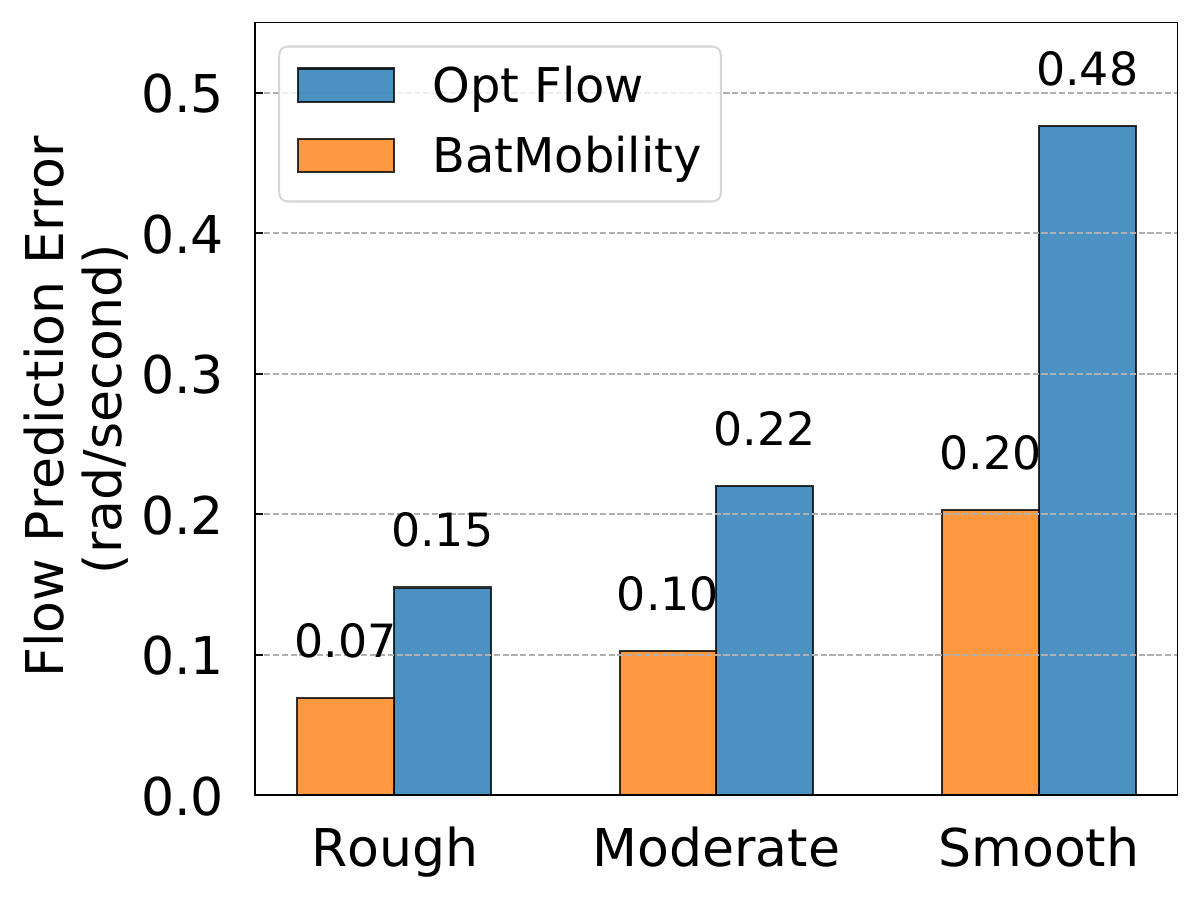}\vspace{-0.1in}
\caption{Accuracy variation across surfaces}
\label{fig:loiter_fps}
\end{subfigure}
\quad
\begin{subfigure}{0.30\textwidth}
\centering
\includegraphics[width=\textwidth]{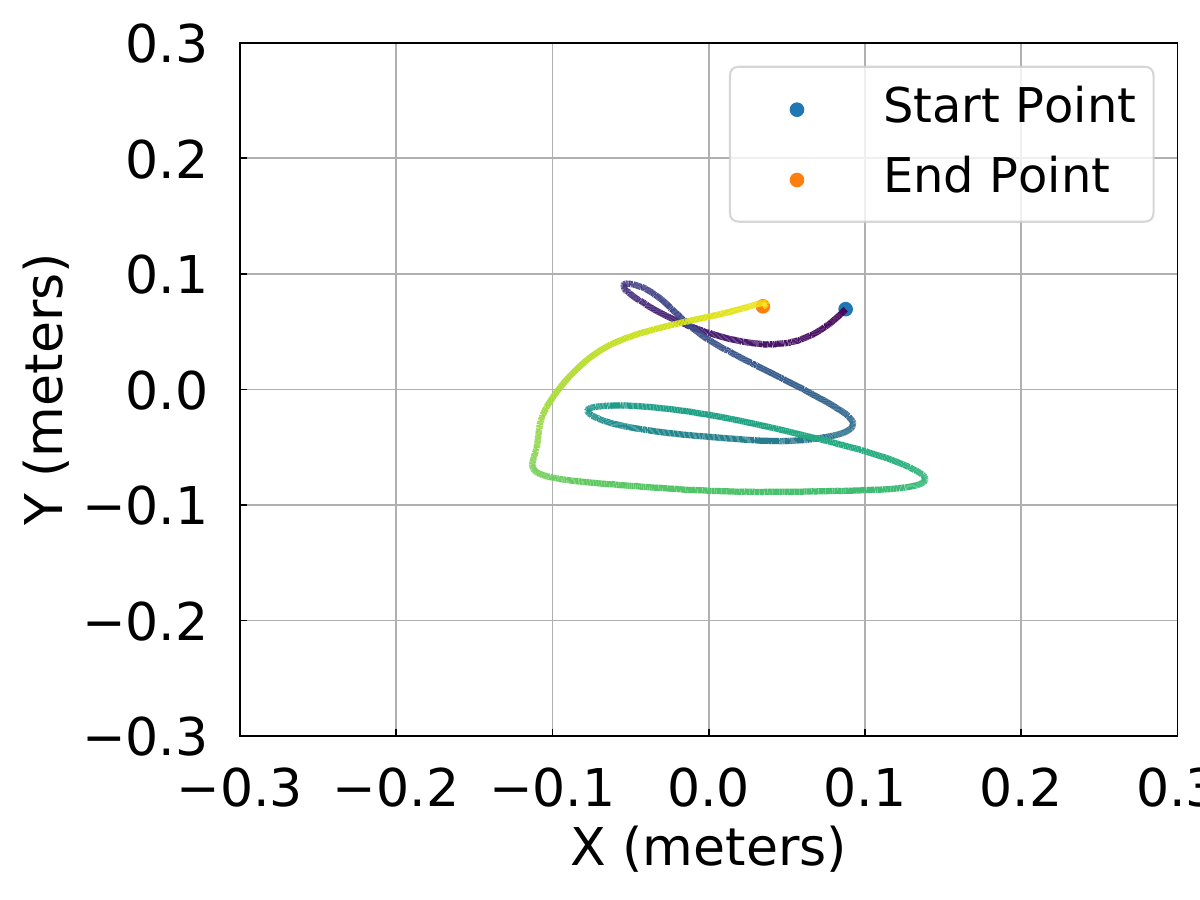}\vspace{-0.1in}
\caption{Qualitative position hold result}
\label{fig:l}
\end{subfigure}

\vspace{-0.1in}
\caption{(a) \name\ outperforms an off-the-shelf optical flow sensor in accurate flow prediction (2.5X) improvement. (b) \name\ outperforms the sensor across surface types, but worsens due to low scattering on smooth surfaces like whiteboards and metal. (c) A UAV can hold its position by responding \red{to motion feedback from \name.} }
\vspace{-0.15in}
\label{fig:flow}
\end{figure*}

\begin{figure*}[htp]
\centering
\includegraphics[width=\textwidth]{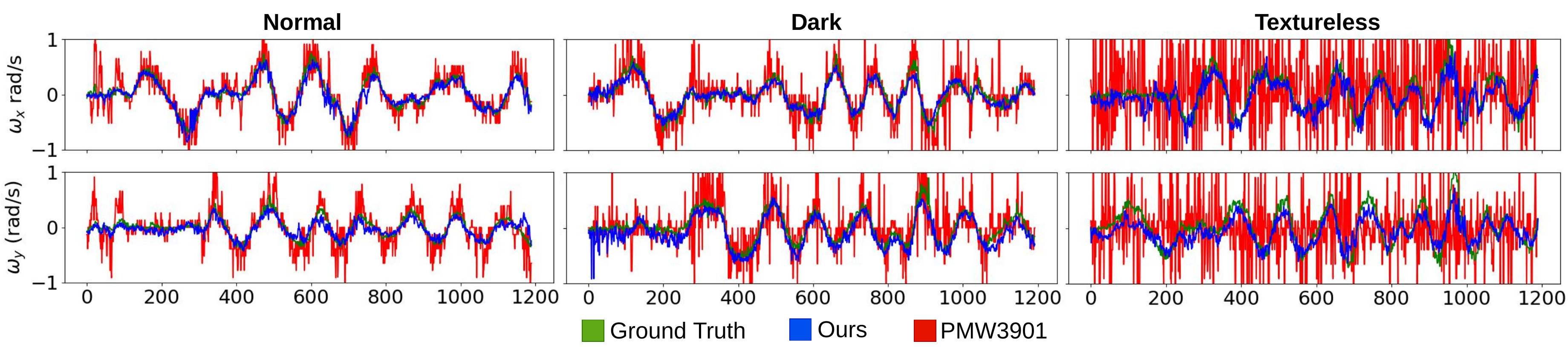}\vspace{-0.15in}
\caption{Sample test traces for our 40 Hz model under three operating conditions on a rough surface. Note that the performance of \name\ is relatively invariant, whereas optical flow struggles under visually adverse conditions.}
\vspace{0.05in}
\label{fig:sample_trace}
\end{figure*}

\subsection{Flow Estimation Accuracy}

\begin{table}
\begin{tabularx}{\columnwidth}{ccccccc}
\toprule
\bf $f_{\text{flow}}$ & \bf $T_{\text{int}}$ & \bf $\theta_{\text{res}}$ & \bf $\theta_{\text{max}}$ & \bf $K$ & $M\times N$ & \bf Model \\
\midrule
15Hz      & 140ms            & $6^\circ$       & $60^\circ$       & 3                        & 24x24        & Mini       \\
30Hz      & 100ms            & $3^\circ$       & $60^\circ$       & 4                        & 24x24        & Micro      \\
40Hz      & 75ms             & $6^\circ$       & $60^\circ$       & 6                        & 24x24        & Micro      \\
50Hz      & 60ms             & $6^\circ$       & $60^\circ$       & 12                       & 12x24        & Nano       \\
\bottomrule
\end{tabularx}
\caption{Preprocessing parameters and model architectures for our 15Hz, 30Hz, 40Hz, and 50Hz models.}\vspace{-0.3in}
\label{tab:parameters}
\end{table}

First, we evaluate \name's flow estimation accuracy under different radar update rate settings and compare with an optical flow sensor as a baseline. For each setting, we collect 12 \red{test} traces in different lighting conditions (dark, and bright), and different textures (rough, moderate, and smooth). \red{These test traces are taken from environments which do not appear in the train/validation dataset.} We extract the flow prediction results from these trajectories and show the overall flow estimation accuracy of different hardware settings in Fig.~\ref{fig:flow}(a). It shows that radar settings with 15, 30, 40Hz always give better predictions than optical flow, and they achieve the median prediction error of 0.091, 0.093 and 0.097 rad/seconds, compared to the optical flow error of 0.230 rad/seconds. This is because optical flow generally does worse on transparent, featureless and dark environments while our radar design is robust to these factors.

To analyze how surface textures affect the flow prediction for both optical flow and \name, we show comparison results in Fig.~\ref{fig:flow}(b) on three different kinds of surfaces: rough (carpet), moderate (floor and table), and smooth (whiteboard and metal). The results show that \name\ achieves %
0.08, 0.12, 0.28 rad/second lower estimation error compared to optical flow in the rough, moderate, and smooth surface respectively. This shows that \name\ equipped with downward facing radar is capable of achieving robust performance in variety of surfaces and this ability is essential to stabilize the UAV \red{during real flight}. Note that the performance of \name\ worsens with the smoothness of the surface because extremely smooth surfaces (like whiteboard) are closer to pure reflectors as opposed to scatterers. Since \name\ uses the scattering effect for doppler-estimation, its performance worsens on these polished surfaces. 

\vspace{-0.1in}\subsection{\red{Position Hold Performance}}
We \red{compare} the position holding (hovering) performance of the UAV in a VICON room \red{when using the optical flow sensor and \name.} These systems \red{are needed to} provide feedback to the UAV to correct drifts in its position. This is a challenging test because \name\ must capture very low velocity motion to avoid drift in position. \red{Futhermore, note that the flight room flooring (foam) is a completely novel surface that was not seen during \name's training.}%

We show qualitative results in Fig.~\ref{fig:flow}(c). The figure shows a 20 seconds trajectory of the UAV equipped with \name\ when we set it to be in the position holding mode. It shows that \name\ manages to stabilize the UAV within the radius of 0.1m by solely using radar. Any drifts in its motion are estimated by \name\ accurately and leads to course correction by the UAV. 

\begin{figure*}[htp]
\centering
\begin{subfigure}{0.30\textwidth}
\centering
\includegraphics[width=\textwidth]{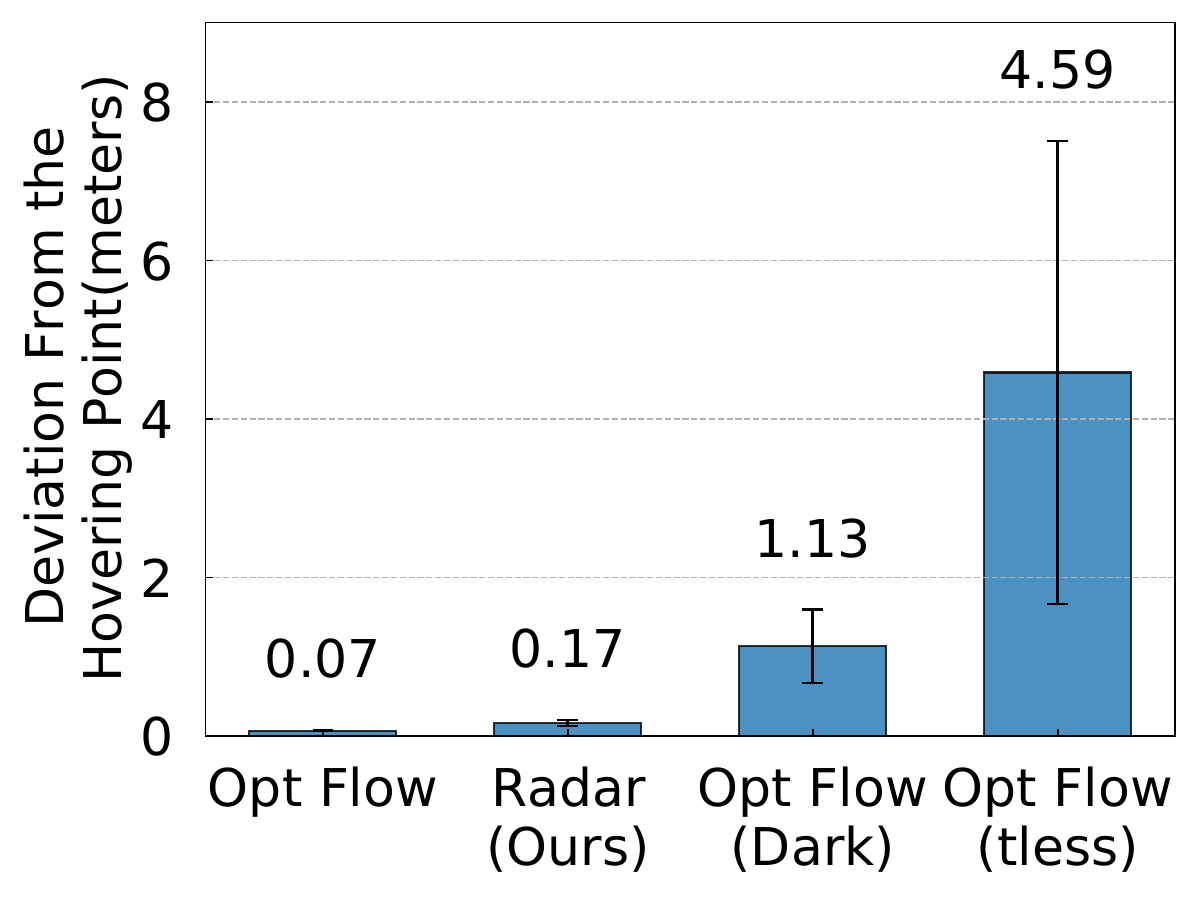}\vspace{-0.1in}
\caption{Median Deviation from Position}
\label{fig:loiter_method}
\end{subfigure}
\quad
\begin{subfigure}{0.30\textwidth}
\centering
\includegraphics[width=\textwidth]{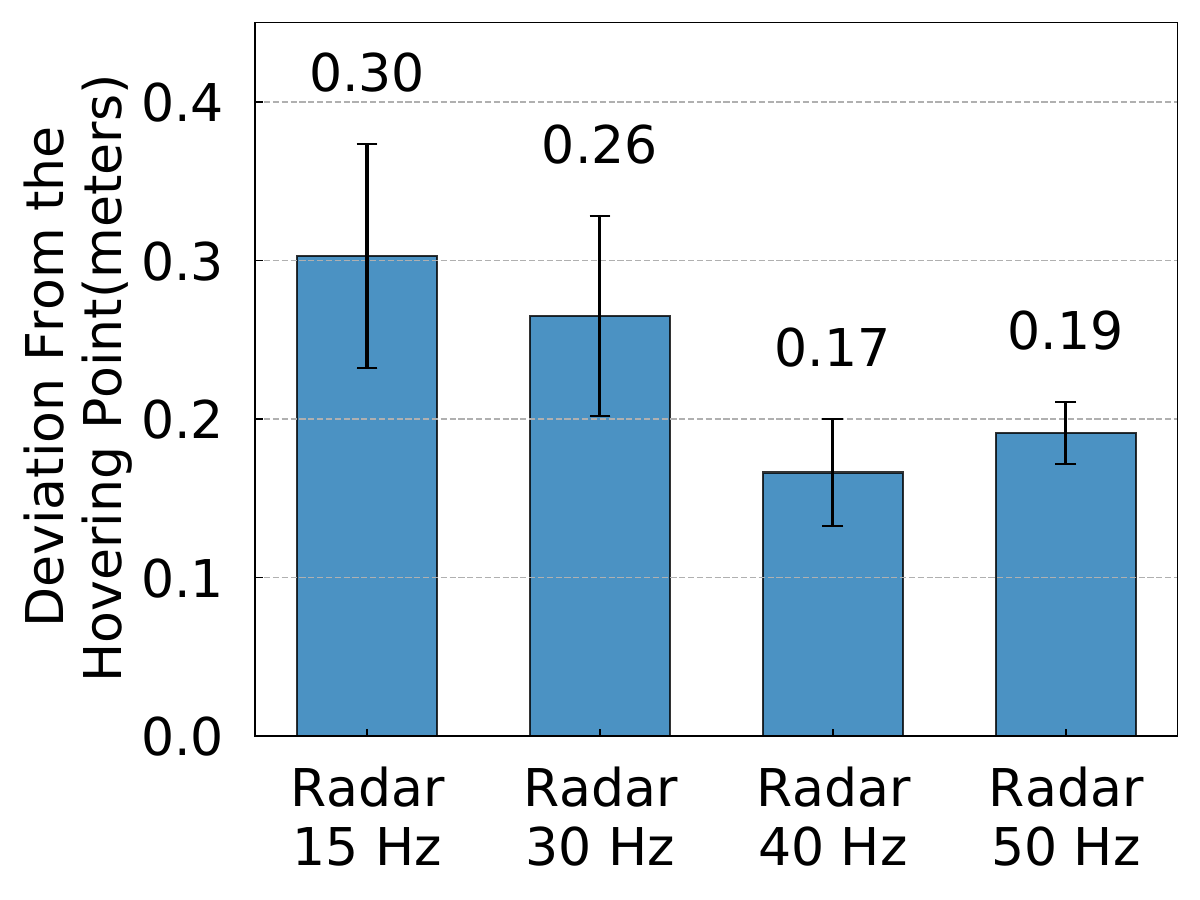}\vspace{-0.1in}
\caption{Deviation vs Update Rate}
\label{fig:loiter_fps}
\end{subfigure}
\quad
\begin{subfigure}{0.30\textwidth}
\centering
\includegraphics[width=\textwidth]{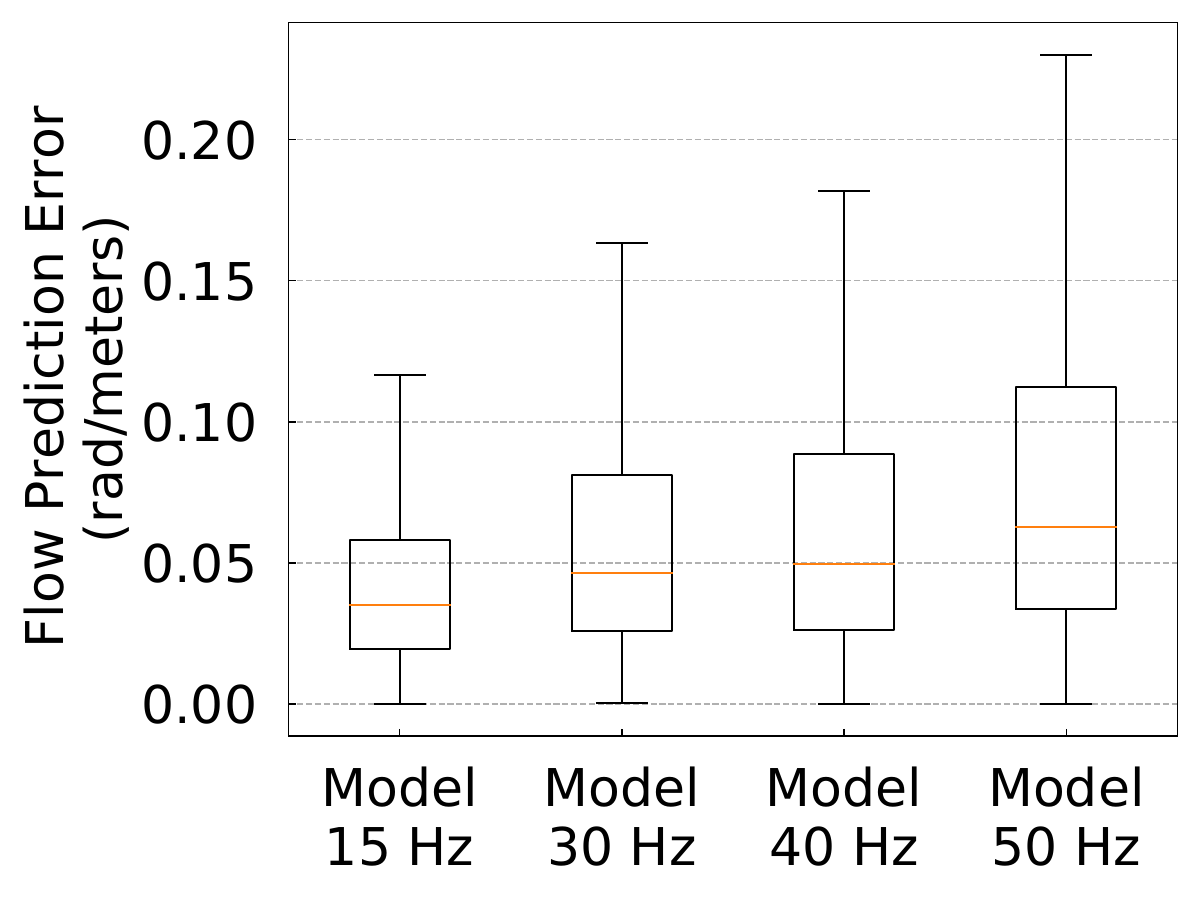}\vspace{-0.1in}
\caption{Flow accuracy vs update rate}
\label{fig:l}
\end{subfigure}
\vspace{-0.1in}
\caption{Loiter Test. (a) UAV equipped with \name\ holds its position, but optical flow fails in dark and textureless conditions. (b) Higher update rates support better hovering performance, in spite of higher flow prediction errors shown in (c).}
\vspace{-0.1in}
\label{fig:loiter}
\end{figure*}

Next, we present quantitative results. We test the optical flow sensor performance in three different environment conditions: the textured surface with sufficient lighting, the dark environment, and the textureless surface. For each setting, we collect 5 trajectories and each trajectory has a duration of 10 seconds. We compare the optical flow sensor with the 40Hz radar pipeline and the results are shown in Figure \ref{fig:loiter}(a). As shown, the UAV equipped with optical flow sensor can hold the position well in the normal environment with the average deviation of 0.07 meters. Note that, optical flow performs better than \name's deviation of 0.19 meters in such settings (despite a worse flow accuracy than \name). This is because the optical flow sensor operates at a higher update rate (at least 50 Hz) and provides more frequent feedback. However, when it comes to the dark and textureless environment, we notice that the optical flow sensor quickly loses the ability to give correct flow prediction and the UAV crashes in less than 10 seconds. %

\begin{figure*}[htp]
\centering
\begin{subfigure}{0.30\textwidth}
\centering
\includegraphics[width=\textwidth]{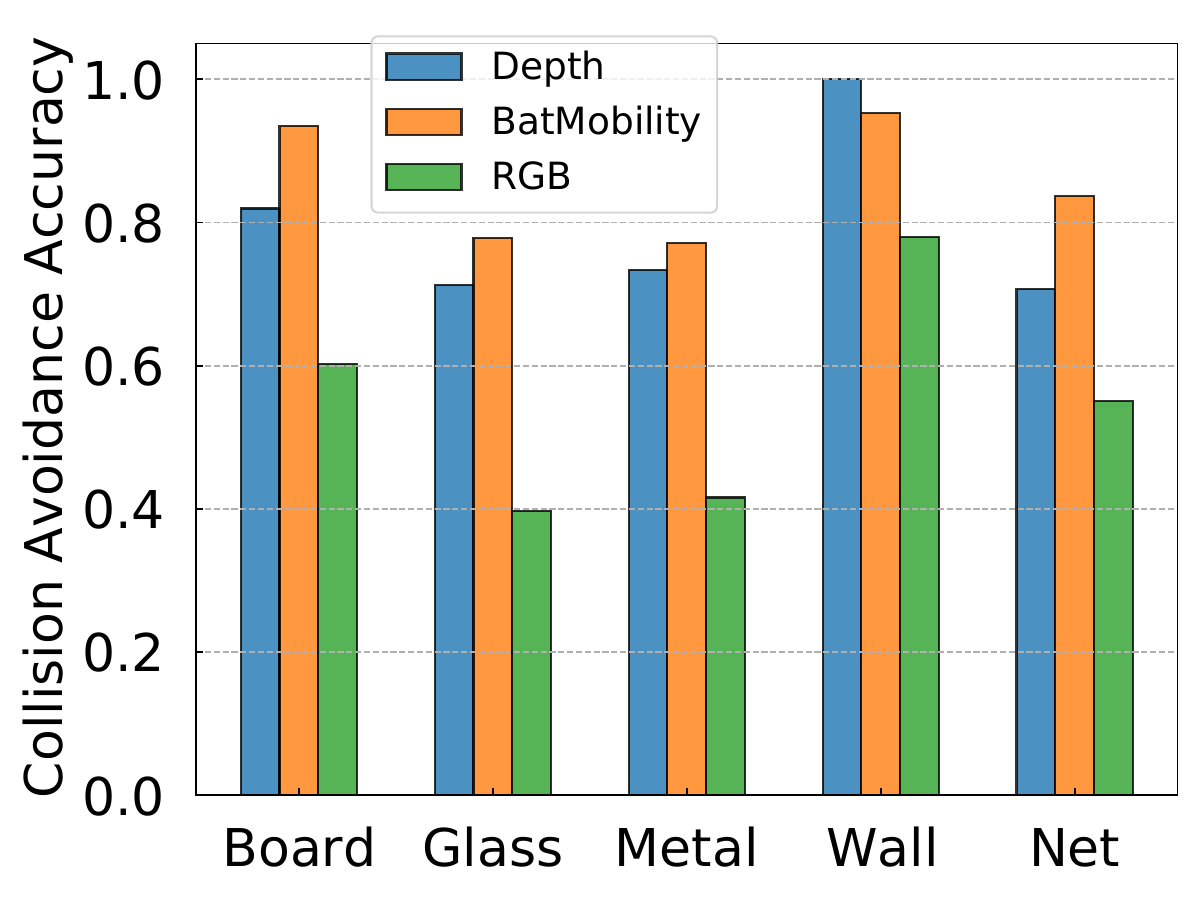}\vspace{-0.1in}
\caption{Accuracy on different materials}
\label{fig:loiter_method}
\end{subfigure}
\quad
\begin{subfigure}{0.30\textwidth}
\centering
\includegraphics[width=0.8\textwidth]{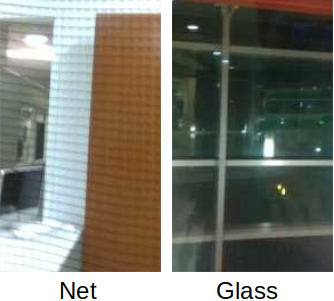}\vspace{-0.1in}
\caption{Examples of RGB collision}
\label{fig:loiter_fps}
\end{subfigure}
\quad
\begin{subfigure}{0.30\textwidth}
\centering
\includegraphics[width=0.8\textwidth]{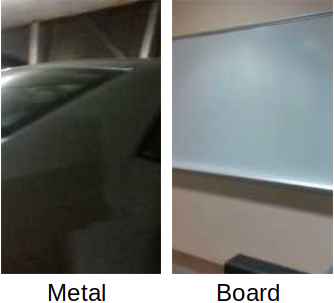}\vspace{-0.1in}
\caption{Examples of Radar collision}
\label{fig:l}
\end{subfigure}

\vspace{-0.1in}
\caption{Collision detection. (a) We test the collision detection accuracy on different materials. \name\ achieves highest accuracy in vast majorities of cases. (b) Two examples where RGB system \red{misses} the obstacle but \name\ succeeds. (c) Two examples of \name\ failure cases on \red{specular} surfaces.}
\vspace{0.05in}
\label{fig:ca}
\end{figure*}

\vspace{-0.1in}\subsection{\red{Update Rate vs Performance}}~\label{sec:tradeoff_empirical}
We investigate the tradeoff between the update rate and hovering performance. We plot the median deviation from intended position in Fig.~\ref{fig:loiter}(b) for different update rates. It shows that when update rate goes up, the deviation from the hovering point gets smaller. This is because more frequent updates allow the flight controller to react sooner to any drift from its position.

However, at 50 Hz, the deviation increases. This is because of two reasons: (a) As shown in Fig.~\ref{fig:loiter}(c), the flow prediction accuracy is lower at higher update rates, and (b) we observe that our software platform is unable to consistently output updates at 50 Hz, thereby dropping some updates. This is a limitation of our current design and is fixable by a more hardware focused design (e.g., using FPGAs). We also note that we compare to a hardware implementation of a commercial optical flow sensor which outputs 50 Hz consistently. Therefore, we stick to 40 Hz as our final design.%

\subsection{\red{Collision Detection}}
We test the \red{collision detection} ability when UAV is facing different materials including whiteboard, glass, metal, wall, and net in our flight room. We compare the collision detection accuracy with two other baseline \red{modalities: depth from an infrared stereo camera, and RGB}. For each surface, we collect 2 minute traces of data and results are shown in Fig.~\ref{fig:ca}. The result shows that \name\ achieves 93\%, 78\%, 77\%, 95\%, and 84\% collision prediction accuracy on the \red{aforementioned} surfaces respectively. We outperform the baselines. Vision-based methods suffer against glass (transparency) and metal (reflective surface). \name's performance is also reduced against glass and metal due to their near specular reflection as opposed to scattering. Note that, these are frame-level collision prediction results. A flight controller can aggregate multiple frames before it decides to stop or move forward. 

We show two examples in Fig.~\ref{fig:ca}(b) where the RGB-aided system \red{misses} the obstacle. This is because \red{nets and glass} are inconspicuous to RGB cameras, especially when the UAV is moving. We also show 2 cases in Fig.~\ref{fig:ca}(c) where \name\ fails to make right decisions. This is because metal and board surfaces are smooth and there is a chance that the transmitted signal undergoes specular reflection and doesn't \red{reflect} back to the receive antenna.

\subsection{Flow Ablation Study}

\begin{table}
\begin{tabularx}{\columnwidth}{ccccccc}
\toprule
\multicolumn{5}{c}{\bf Parameters} & \multirow{2}{*}{\bf ResNet18} & \multirow{2}{*}{\bf Mini} \\
\cmidrule(lr){1-5}
$T_{\text{int}}$ & $\theta_{\text{res}}$ & $\theta_{\text{max}}$ & $K$ & $M,N$ & & \\ \midrule
70ms  & $4^\circ$ & $90^\circ$ & $3$ & $32, 32$ & 0.071 & 0.076 \\
140ms & $4^\circ$ & $90^\circ$ & $3$ & $32, 32$ & 0.069 & 0.071 \\
140ms & $6^\circ$ & $60^\circ$ & $3$ & $32, 32$ & 0.071 & 0.072 \\ 
140ms & $6^\circ$ & $60^\circ$ & $4$ & $32, 32$ & 0.069 & 0.073 \\
140ms & $6^\circ$ & $60^\circ$ & $4$ & $24, 24$ & 0.071 & 0.074 \\
\bottomrule
\end{tabularx}
\caption{Effect of preprocessing parameters and model architecture on flow validation RMSE (rad/s). Lower is better.}
\label{tab:ablation}
\vspace{-0.3in}
\end{table}

What effect does varying the preprocessing parameters and choice of model architecture have on the flow prediction performance? We retrain the 15 Hz model with different parameters and report the validation accuracy in Table \ref{tab:ablation}. We find that the two biggest determinants are the choice of model architecture \red{(i.e. ResNet18, ResNet18Mini)} and the \red{integration time $T_{\text{int}}$}. By contrast, the effect of varying preprocessing parameters \red{($\theta_{\text{res}}, \theta_{\text{max}}, K, M, N)$} on the flow accuracy is usually marginal, yet it can \red{greatly reduce} the pipeline runtime.

\section{Related Work}

\para{RF-Based Odometry:} 
There is a large body of work on the use of spinning radar for ego-motion estimation of cars \cite{kellner_instantaneous_2014,rapp_probabilistic_2017,cen_precise_2018,park_pharao_2020,monaco_radarodo_2020}. Although such radars have exceptional range and angular resolution, their bulkiness, power consumption, and mechanical nature make them unsuitable for small UAVs.

Recent work has focused on bringing radar-based odometry techniques to more compact single-chip mmWave radar sensors. milliEgo \cite{lu_milliego_2020} is a learning-based method that fuses radar point clouds with IMU data. \cite{kramer_radar-inertial_2020} fuses doppler shift from radar point clouds with IMU readings using a factor graph. \cite{park_3d_2021} extends this technique to 3D by leveraging two orthogonally placed radars in addition to an IMU. All such approaches use outward facing radars and thus rely on the presence of static reflectors in the surrounding environment. By contrast, \name's odometry relies solely on a single ground facing radar and does not require an auxiliary IMU. Furthermore, such works are tested only on ground robots, whereas \name~ addresses the challenges of integrating radar-based odometry in a plug-and-play fashion on a real UAV.
 
Finally, another body of work uses wireless localization techniques to provide odometry estimates to UAVs \cite{ledergerber_robot_2015,tiemann_scalable_2017, zhao_3d_2021,chi_wi-drone_2022}. These approaches require pre-existing wireless anchor points in the environment. \name~  does not require such infrastructure and instead relies purely on on-board sensing, thus enabling true autonomy.

\para{RF-Based Obstacle Detection:}
In an automotive context, mmWave radar is often used to detect and image surrounding cars \cite{nabati_rrpn_2019,guan_through_2020, bansal_pointillism_2020}. Analogously, various works propose the use of mmWave radar on UAVs to detect various obstacles that it might crash into \cite{shi_multichannel_2013,dogru_pursuing_2020,wessendorp_obstacle_2021,safa_low-complexity_2021}. We differ from prior works in two ways. Firstly, we do not assume prior knowledge on the kind of obstacle we wish to detect (human, car, another UAV) and instead opt for a generic notion of an obstacle as being anything which a UAV can physically collide. Secondly, unlike prior approaches which operate on preprocessed point clouds, we mindfully opt for an end-to-end approach based on radar heatmaps to prevent any task-relevant cues from being lost during the point cloud conversion process.

\para{Control Policies for UAVs:} Strategies for collision-free flight for UAVs fall under two main categories. The first comprises classical approaches based on planning paths in metric maps of the environment. These works are highly reliant on accurate geometric maps of the environment, which are obtained through relatively expensive optical depth sensors (i.e. stereo depth cameras, lidars). Such approaches include \cite{usenko_real-time_2017,lin_robust_2020,zhou_ego-planner_2021}. The alternative paradigm is learning-based approaches that map raw sensor data to actions \cite{levine_end--end_2016}. For example, vision-based approaches have become popular in recent years due to the ubiquity of cameras and better means of generating image data at scale. Work in this space includes \cite{ross_learning_2013,sadeghi_cad2rl_2017,gandhi_learning_2017,kaufmann_deep_2018,bonatti_learning_2020}. We note that BatMobility is designed with this paradigm in mind.
\section{Concluding Discussion}

We present \name, a learning-based pure radar perception system that enables autonomous drones to fly in unstructured and unknown environments. We highlight the advantages and disadvantages of an RF-only approach when compared to a conventional optical sensor based system. The core of our approach is a novel radio flow primitive that can be broadly useful in other contexts involving motion sensing, As such, we hope this work opens up new frontiers in the use of RF sensing for robotics, self-driving vehicles, VR/AR, and beyond. We conclude with a brief discussion of limitations and future work:

\para{Deploying in External Environments: }We test \name\ in a controlled flight room environment to measure its performance with VICON cameras. However, our evidence suggests that the system would work just as well or even better outdoors. This is because our experiments were performed on a foam and smooth concrete floor, which appear as highly smooth surfaces to the radar. As our results show, \name\ works better on rough surfaces, which are ubiquitous in outdoor environments (e.g. grass, asphalt roads). 

\para{System Constraints: } Our system is primarily constrained by the maximum operating range and maximum unambiguous velocity of the radar. These are determined by the radar chirp parameters as explained in Sec.~\ref{sec:background} and \ref{sec:design}. For our evaluation, we choose chirps with maximum range within 3-5 m and maximum velocity of 1.5-2 m/s. We believe this is appropriate for \name\'s ideal use case, i.e. flying cautiously at low altitudes in order to prevent collisions with surrounding objects. One potential avenue of future exploration is to adaptively tune the radar's chirp parameters depending on the current scenario (i.e. increasing maximum range at higher altitudes, increasing maximum velocity when moving faster, etc).

\para{Achieving Agile Flight: }We focus only on 2-D action primitives in \name\ (i.e. moving forward, turning, etc). Although such an action set is practical for navigation purposes, it does not encompass the full set of a quadcopter's capabilities. This prevents us from, for example, performing drone acrobatics \cite{kaufmann_deep_2020}, or following time-optimal trajectories through environments \cite{foehn_alphapilot_2020}. We leave such exploration to future work.

\vspace{6pt}\noindent\textbf{Acknowledgments --- }We thank the reviewers and our anonymous shepherd for their insightful comments and suggestions on improving this paper. This work was supported in part by NSF RINGS Award 2148583. This work was carried out in part in the Intelligent Robotics Laboratory, University of Illinois Urbana-Champaign. We thank John M. Hart for help regarding the flying arena. We thank Shahab Nikkhoo for his guidance and suggestions. We thank Kris Hauser for letting us use his 3D printer. We are grateful to Jayanth Shenoy, Bill Tao, Maleeha Masood, Ishani Janveja, and Om Chabra for their feedback on initial drafts.

\newpage

\bibliographystyle{ACM-Reference-Format}
\bibliography{acmart}

\end{document}